\RequirePackage{fix-cm}
\documentclass[twocolumn]{svjour3}           %
\smartqed  %
\usepackage{graphicx}
\usepackage{amsmath,amssymb} %
\usepackage{soul}
\usepackage{subfig}
\usepackage[dvipsnames]{xcolor}
\usepackage{tabularx} %
\usepackage[pagebackref=true,breaklinks=true,colorlinks,bookmarks=false]{hyperref}
\usepackage{booktabs}
\usepackage{pifont}
\usepackage{bold-extra}
\usepackage{dirtytalk}
\usepackage{mathtools}
\usepackage[numbers,sort]{natbib} %

\usepackage[normalem]{ulem}
\newcommand{\cmark}{\ding{51}}%
\newcommand{\xmark}{\ding{55}}%
\newcommand{\bsldict}{\textsc{BslDict}}
\newcommand{\bslonek}{\text{BSL-1K}}
\newcommand{\baselineDict}{$\text{I3D}^{\text{\bsldict}}$}
\newcommand{\baselineBbc}{$\text{I3D}^{\text{\bslonek}}$}
\newcommand{\baselineBbcDict}{$\text{I3D}^{\text{\bslonek},\text{\bsldict}}$}
\newcommand{\trainspot}{Train$^{\text{ReT}}$}
\newcommand{\testspot}{Test$^{\text{ReT}}$}
\newcommand{\trainrec}{Train$^{\text{Rec}}$}
\newcommand{\testrec}{Test$^{\text{Rec}}_{2K}$}
\newcommand{\testrecnew}{Test$^{\text{Rec}}_{37K}$}

\def\sepappendix{0} %
\usepackage{multirow}
\usepackage{siunitx}
\begin{document}

\title{Scaling up sign spotting through sign language dictionaries} %

\author{G\"ul Varol$^{1,2*}$\thanks{*Equal contribution} \and Liliane Momeni$^{1*}$ \and Samuel Albanie$^{1,3*}$ \\ Triantafyllos Afouras$^{1}$ \and Andrew Zisserman$^1$}
\authorrunning{Varol, Momeni, Albanie, Afouras, Zisserman}

\institute{
    $^{1}$ Visual Geometry Group, University of Oxford, UK \\
    $^{2}$ LIGM, \'Ecole des Ponts, Univ Gustave Eiffel, CNRS, France \\
    $^{3}$ Department of Engineering, University of Cambridge, UK \\
    \email{\tt\small \{gul,liliane,albanie,afourast,az\}@robots.ox.ac.uk} \\
    {\tt\small \url{https://www.robots.ox.ac.uk/~vgg/research/bsldict/} }
}

\date{Received: 1 May 2021 / Accepted: 21 January 2022}

\maketitle

\begin{abstract} 
The focus of this work is \textit{sign spotting}---given a video of an isolated sign, our task is to identify \textit{whether} and \textit{where} it has been signed in a continuous, co-articulated sign language video. To achieve this sign spotting task, we train a model using multiple types of available supervision by: (1) \textit{watching} existing footage which is sparsely labelled using mouthing cues; (2) \textit{reading} associated subtitles (readily available translations of the signed content) which provide additional \textit{weak-supervision}; (3) \textit{looking up} words (for which no co-articulated labelled examples are available) in visual sign language dictionaries to enable novel sign spotting.
These three tasks are integrated into a unified learning framework using the principles of Noise Contrastive Estimation and Multiple Instance Learning.  We validate the effectiveness of our approach on low-shot sign spotting benchmarks. 
In addition, we contribute a machine-readable British Sign Language (BSL) dictionary dataset of isolated signs, \bsldict, to facilitate study of this task.
The dataset, models and code are available at our project page.

\end{abstract}
\section{Introduction} \label{sec:intro}

The objective of this work is to develop a \textit{sign spotting} model that can identify 
and localise instances of signs
within sequences of continuous sign language. Sign languages represent the natural means of 
communication for deaf communities~\cite{sutton-spence_woll_1999} and sign spotting has a 
broad range of practical applications. Examples include: indexing videos of signing content 
by keyword to enable content-based search; gathering diverse dictionaries of sign exemplars 
from unlabelled footage for linguistic study; automatic feedback for language students via 
an \say{auto-correct} tool (e.g. \say{did you mean this sign?}); making voice activated 
wake word devices
available
to deaf communities; and building sign language datasets by 
automatically labelling examples of signs.

Recently, deep neural networks, equipped with large-scale, labelled datasets
produced considerable progress in audio~\cite{coucke2019efficient,veniat2019stochastic} and 
visual~\cite{Momeni20,stafylakis2018zero} keyword spotting in \textit{spoken languages}. 
However, a direct replication of these keyword spotting successes in sign language requires 
a commensurate quantity of labelled data (note that modern audiovisual spoken keyword 
spotting datasets contain millions of densely labelled 
examples~\cite{chung2017lip,afouras2018lrs3}), but such datasets are not available for sign language.

\begin{figure*}[t]
    \centering
    \includegraphics[width=\linewidth]{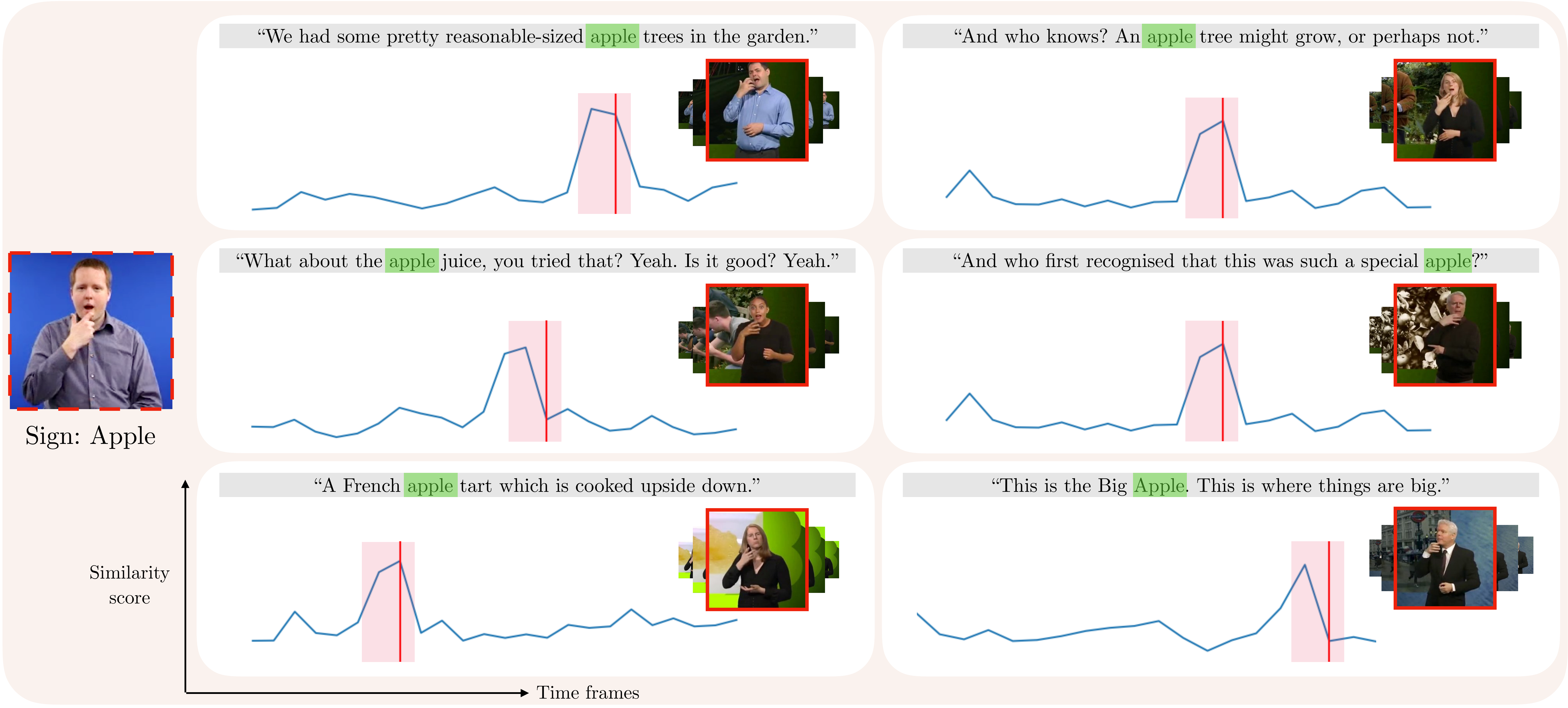}
    \caption{
    We consider the task of \textit{sign spotting} in co-articulated, continuous signing. Given a
    query dictionary video of
    an isolated sign (e.g., ``apple''),
    we aim to identify \textit{whether} and \textit{where} it appears in videos of continuous signing.
    The wide domain gap between dictionary examples of \textit{isolated} signs and target sequences of \textit{continuous} signing makes the task extremely challenging.
    }
    \label{fig:kws-task}
\end{figure*}

It might be thought that a sign language dictionary would offer a relatively 
straightforward solution to
the sign spotting task, particularly to the problem of covering only a limited vocabulary 
in existing large-scale corpora.
But, unfortunately, this is not the case due to the severe \textit{domain differences} 
between dictionaries 
and continuous signing in the wild.
The challenges are that sign language dictionaries typically: (1)
consist of \textit{isolated signs} which differ in appearance from the
\textit{co-articulated}\footnote{\textit{Co-articulation} refers to changes in the appearance of the 
current sign due to neighbouring signs.} sequences of continuous signs (for which we ultimately
wish to perform spotting); and  (2) differ in speed (are performed more slowly) relative to
co-articulated signing. 
Furthermore, (3) dictionaries only  possess a few examples of each sign (so learning must be \textit{low shot});
and as one more challenge, (4) there can be multiple signs corresponding to a single keyword, for example
due to regional variations of the sign language~\cite{bslcorpus17}.
We show through experiments in
Sec.~\ref{sec:experiments}, that directly training a sign spotter for
continuous signing on dictionary examples,
obtained from an internet-sourced sign
language dictionary, does indeed perform poorly.

To address these challenges, we 
propose
a unified framework in which sign spotting embeddings are learned
from the dictionary (to provide broad coverage of the lexicon) in combination with two 
additional sources of supervision.  
In aggregate, these multiple types of supervision include: (1)~\textit{watching} sign language and learning from existing sparse
annotations obtained from mouthing cues \cite{Albanie20}; (2)~exploiting
weak-supervision by \textit{reading} the subtitles that accompany the
footage and extracting candidates for signs that we expect to be
present; (3)~\textit{looking up} words (for which we do not have
labelled examples) in a sign language dictionary. The recent development of a large-scale, 
subtitled dataset
of continuous signing providing sparse annotations~\cite{Albanie20} 
allows us to study this problem setting directly.
We formulate our
approach as a Multiple Instance Learning problem in which positive
samples may arise from any of the three sources and employ Noise
Contrastive Estimation~\cite{gutmann2010noise} to learn a domain-invariant (valid across 
both isolated and co-articulated signing)
representation of signing content.

Our loss formulation is an extension of InfoNCE \cite{oord2018representation,wu2018unsupervised}
    (and in particular the multiple instance variant \mbox{MIL-NCE} \cite{miech2020end}).
    The novelty of our method lies in the \textit{batch formulation} that leverages
    the mouthing annotations, subtitles, and visual dictionaries to define positive
    and negative bags. Moreover, this work specifically focuses on computing similarities
    across two different domains to learn matching between isolated and co-articulated signing.

We make the following %
contributions, originally introduced in~\cite{momeni20watchread}: (1)~We 
provide a machine readable British Sign Language (BSL) dictionary dataset of isolated 
signs, \bsldict, to facilitate study of the sign spotting task; (2)~We propose a unified 
Multiple Instance Learning framework for learning sign embeddings suitable for spotting 
from three supervisory sources; (3)~We validate the effectiveness of our approach on a 
co-articulated sign spotting benchmark for which only a small number (low-shot) of isolated 
signs are provided as labelled training examples, and (4)~achieve state-of-the-art 
performance on the \bslonek{} sign spotting benchmark~\cite{Albanie20} (closed vocabulary). 
We show qualitatively that the learned embeddings can be used to (5)~automatically mine new 
signing examples, and (6)~discover \say{faux amis} (false friends) between sign languages.
In addition, we extend these contributions with (7)~the demonstration that our framework can be effectively
deployed to obtain large numbers of
sign examples, enabling state-of-the-art performance to be reached on the 
BSL-1K sign recognition benchmark \cite{Albanie20},
and on the recently released BOBSL dataset~\cite{Albanie2021bobsl}.

%Our extensions over the orginal ACCV conference version of this work~\cite{momeni20watchread} include:
%(i) a new section (Sec.~\ref{subsection:visual-kws}) that describes the method of using mouthings for sign spotting; 
%(ii) a new section (Sec.~\ref{subsection:recognition}) that presents additional results on sign language recognition
%(Tab.~\ref{tab:recognitionablation} and Fig.~\ref{fig:stats} are new, Tab.~\ref{tab:recognition} is from our followup work \cite{varol21bslattend});
%and (iii) two additional ablations in Sec.~\ref{subsection:ablations} 
%(Tab.~\ref{tab:mouthingthres} and Tab.~\ref{tab:stride} were previously in the supplementary material);
%(iv) a new section (Sec.~\ref{subsection:bobsl}) that gives results on a new
%publicly available
%dataset: BOBSL;
%(v) a new section (Sec.~\ref{subsection:limitations})
%that discusses limitations.

\section{Related Work}
\label{sec:related}

Our work relates to several themes in the literature: \textit{sign
language recognition} (and more specifically \textit{sign
spotting}), \textit{sign language datasets}, \textit{multiple
instance learning} and \textit{low-shot action localization}. We discuss each of these themes next. %

\noindent \textbf{Sign language recognition.}  
The study of automatic sign recognition has a rich history in the
computer vision community stretching back over 30 years, with early
methods developing carefully engineered features to model trajectories
and shape~\cite{Kadir04a,Tamura88,Starner95,Fillbrandt2003}. A series of
techniques then emerged which made effective use of hand and body pose
cues through robust keypoint estimation
encodings~\cite{Buehler09,Cooper2011,Ong2012,Pfister14}.  Sign
language recognition also has been considered in the context of
sequence prediction, with
HMMs~\cite{Ulrich2008,Forster2013,Starner95,Kadir04a},
LSTMs~\cite{Camgoz17,Huang2018VideobasedSL,Ye18,zhou2020spatialtemporal},
and Transformers~\cite{camgoz2020sign} proving to be effective mechanisms
for this task. Recently, convolutional neural networks have emerged as
the dominant approach for appearance modelling~\cite{Camgoz17}, and in
particular, action recognition
models using spatio-temporal convolutions~\cite{Carreira2017} have proven very well-suited for
video-based sign
recognition~\cite{Joze19msasl,Li19wlasl,Albanie20}. We adopt the I3D
architecture~\cite{Carreira2017} as a foundational building block in our studies.

\noindent\textbf{Sign language spotting.}
The sign language spotting problem---in which the objective is to find performances of a 
sign (or sign sequence) in a longer sequence of signing---has been studied with Dynamic 
Time Warping and skin colour histograms~\cite{viitaniemi14} and with Hierarchical 
Sequential Patterns~\cite{ong2014}. Different from our work which learns representations 
from multiple weak supervisory cues, these approaches consider a fully-supervised setting 
with a single source of supervision and use hand-crafted features to represent signs~\cite{Farhadi07}. 
Our proposed use of a dictionary is also closely tied to \textit{one-shot/few-shot 
learning}, in which the learner is assumed to have access to only a handful of annotated 
examples of the target category. One-shot dictionary learning was studied 
by~\cite{Pfister14} -- different to their approach, we explicitly account for
variations in the dictionary for a given word
(and validate the improvements brought by doing so in 
Sec.~\ref{sec:experiments}).  Textual descriptions from a dictionary of 250 signs were used 
to study zero-shot learning by~\cite{Bilge19ZS} -- we instead consider the practical 
setting in which a handful of video examples are available per-sign and work with a
much larger vocabulary (9K words
and phrases). %

The use of dictionaries to locate signs in subtitled video also 
shares commonalities with \textit{domain adaptation}, since our method must bridge 
differences between the dictionary and the target continuous signing distribution. A vast 
number of techniques have been proposed to tackle distribution shift, including several 
adversarial feature alignment methods that are specialised for the few-shot 
setting~\cite{Motiian2017FewShotAD,Zhang2019}.  
In our work, we explore the domain-specific batch normalization (DSBN) method 
of~\cite{chang2019domain}, finding ultimately that simple batch normalization parameter 
re-initialization is instead most effective when jointly training on two domains after pre-training 
on the bigger domain. The concurrent work of~\cite{li2020transferring} also seeks to align 
representation of isolated and continuous signs. 
However, our work differs from theirs in several key aspects: (1) rather than assuming 
access to a large-scale labelled dataset of isolated signs, we consider the setting in 
which only a handful of dictionary examples may be used to represent a word; (2) we develop 
a generalised Multiple Instance Learning framework which allows the learning of 
representations from weakly-aligned subtitles whilst exploiting sparse labels from 
mouthings \cite{Albanie20} and dictionaries (this integrates cues beyond the learning 
formulation in~\cite{li2020transferring}); (3) we seek to label and improve performance on 
co-articulated signing (rather than improving recognition performance on isolated signing). 
Also related to our work, \cite{Pfister14} uses a \say{reservoir} of weakly labelled sign 
footage to improve the performance of a sign classifier learned from a small number of 
examples. Different to~\cite{Pfister14}, we propose a multiple instance learning formulation 
that explicitly accounts for signing variations that are present in the dictionary.

\begin{figure*}[t]
    \centering
    \includegraphics[width=0.9\textwidth]{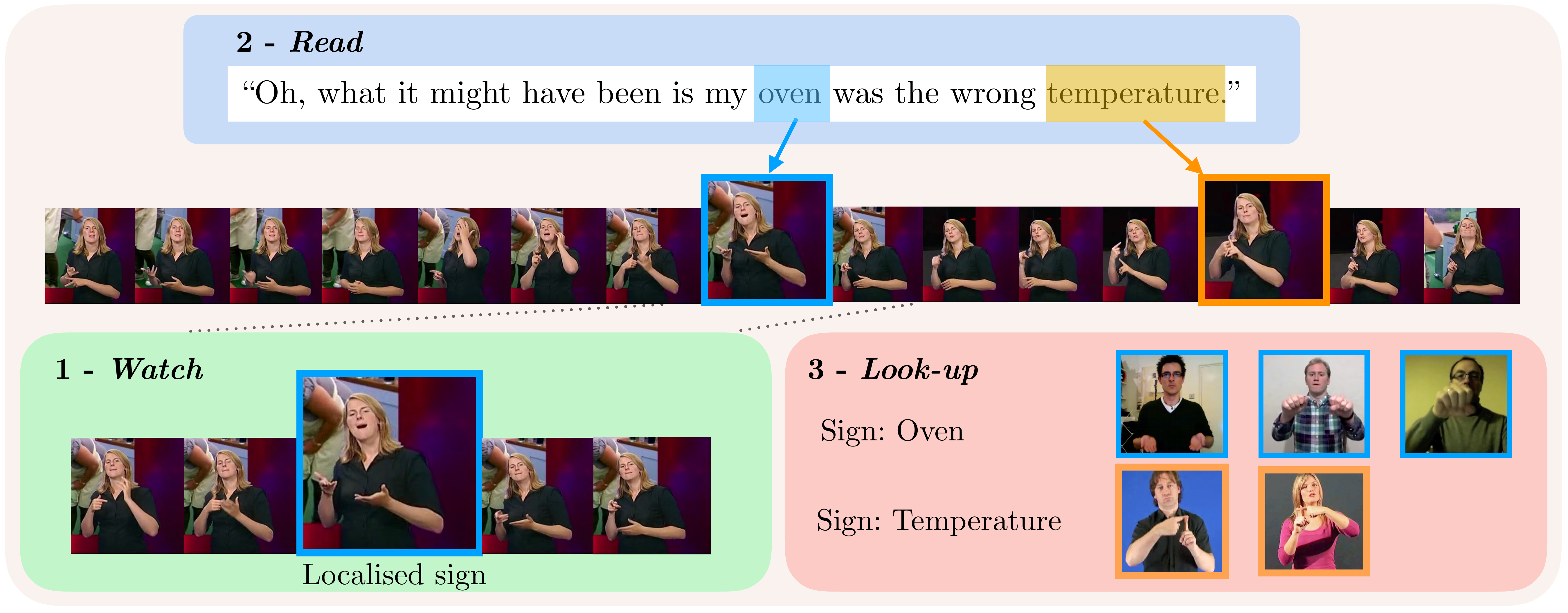}
    \caption{\textbf{The proposed \textit{Watch, Read and Lookup} framework} trains sign spotting embeddings with three cues: (1) \textit{watching} videos and learning from sparse annotation in the form of localised signs obtained from mouthings \cite{Albanie20} (lower-left); (2) \textit{reading} subtitles to find candidate signs that may appear in the source footage (top); (3) \textit{looking up} corresponding visual examples in a sign language dictionary and aligning the representation against the embedded source segment (lower-right).}
    \label{fig:watch-read-lookup}
\end{figure*}

\noindent\textbf{Sign language datasets.} A number of sign language datasets have been 
proposed for studying Finnish~\cite{viitaniemi14}, German~\cite{Koller15cslr,signum2008}, 
American~\cite{asllvid2008,Joze19msasl,Li19wlasl,purdue06} and 
Chinese~\cite{chai2014devisign,Huang2018VideobasedSL} sign recognition. For British Sign 
Language (BSL), \cite{schembri2013building} gathered the BSL Corpus which
represents continuous signing, labelled with %
fine-grained linguistic annotations.
More recently~\cite{Albanie20} collected BSL-1K, a 
large-scale dataset of BSL signs that were obtained using a mouthing-based keyword spotting 
model. Further details on this method are given in Sec.~\ref{subsection:visual-kws}. 
In this work, we contribute \mbox{\bsldict{}}, a dictionary-style dataset that is complementary to 
the datasets of~\cite{schembri2013building,Albanie20} -- it contains only a handful of 
instances of each sign, but achieves a comprehensive coverage of the BSL lexicon with a 9K 
English vocabulary (vs a 1K vocabulary in~\cite{Albanie20}). As we show in the sequel, this dataset 
enables a number of sign spotting applications.
While \bsldict{} does not represent a linguistic corpus,
as the correspondences to English words and phrases are not carefully annotated
with \textit{glosses}\footnote{Glosses are atomic lexical units used to annotate sign languages.},
it is significantly larger than its linguistic counterparts (e.g., 4K videos
corresponding to 2K words in BSL SignBank \cite{signbank2014}, as opposed
to 14K videos of 9K words in \bsldict{}),
therefore \bsldict{} is particularly suitable to be used in conjunction with subtitles.

\noindent\textbf{Multiple instance learning.} Motivated by the readily available sign 
language footage that is accompanied by subtitles, a number of methods have been proposed 
for learning the association between signs and words that occur in the subtitle text 
\cite{Buehler09,Cooper2009,Pfister14,Chung16b}.  In this work, we adopt the framework of 
Multiple Instance Learning (MIL)~\cite{dietterich1997solving} to tackle this problem, 
previously explored by~\cite{Buehler09,pfister2013large}. Our work differs from these works 
through the incorporation of a dictionary, and a principled mechanism for explicitly 
handling sign variants, to guide the learning process. Furthermore, we generalise the MIL 
framework so that it can learn to further exploit sparse labels. We also conduct 
experiments at significantly greater scale to make use of the full potential of MIL, 
considering more than two orders of magnitude more weakly supervised data 
than~\cite{Buehler09,pfister2013large}.

\noindent\textbf{Low-shot action localization.} 
This theme investigates semantic video localization: given one or more query videos the 
objective is to localize the segment in an untrimmed video that corresponds  semantically 
to the query video~\cite{Feng_2018_ECCV,Yang_2018_CVPR,Cao_2020_CVPR}.
Semantic matching is too general for the sign-spotting considered in this paper. However, 
we build on the temporal ordering ideas explored in this theme.

\section{Learning Sign Spotting Embeddings from Multiple Supervisors} \label{sec:method}

In this section, we describe the task of \textit{sign spotting} and the three forms of 
supervision we assume access to. Let $\mathcal{X}_{\mathfrak{L}}$ denote the space of RGB 
video segments containing a frontal-facing individual communicating in sign language 
$\mathfrak{L}$ and denote by $\mathcal{X}_{\mathfrak{L}}^{\text{single}}$ its restriction 
to the set of segments containing a single sign. Further, let $\mathcal{T}$ denote the 
space of subtitle sentences and $\mathcal{V}_{\mathfrak{L}}=~\{1, \dots, V \}$ denote the 
\textit{vocabulary}---an index set corresponding to an enumeration of written words that 
are equivalent to signs that can be performed in $\mathfrak{L}$\footnote{Sign language 
dictionaries provide a word-level or phrase-level correspondence
(between sign language and spoken language) for many signs but no universally accepted 
\textit{glossing} scheme exists for transcribing languages such as 
BSL~\cite{sutton-spence_woll_1999}.}.

\begin{figure*}[t]
    \centering
    \includegraphics[width=0.9\textwidth]{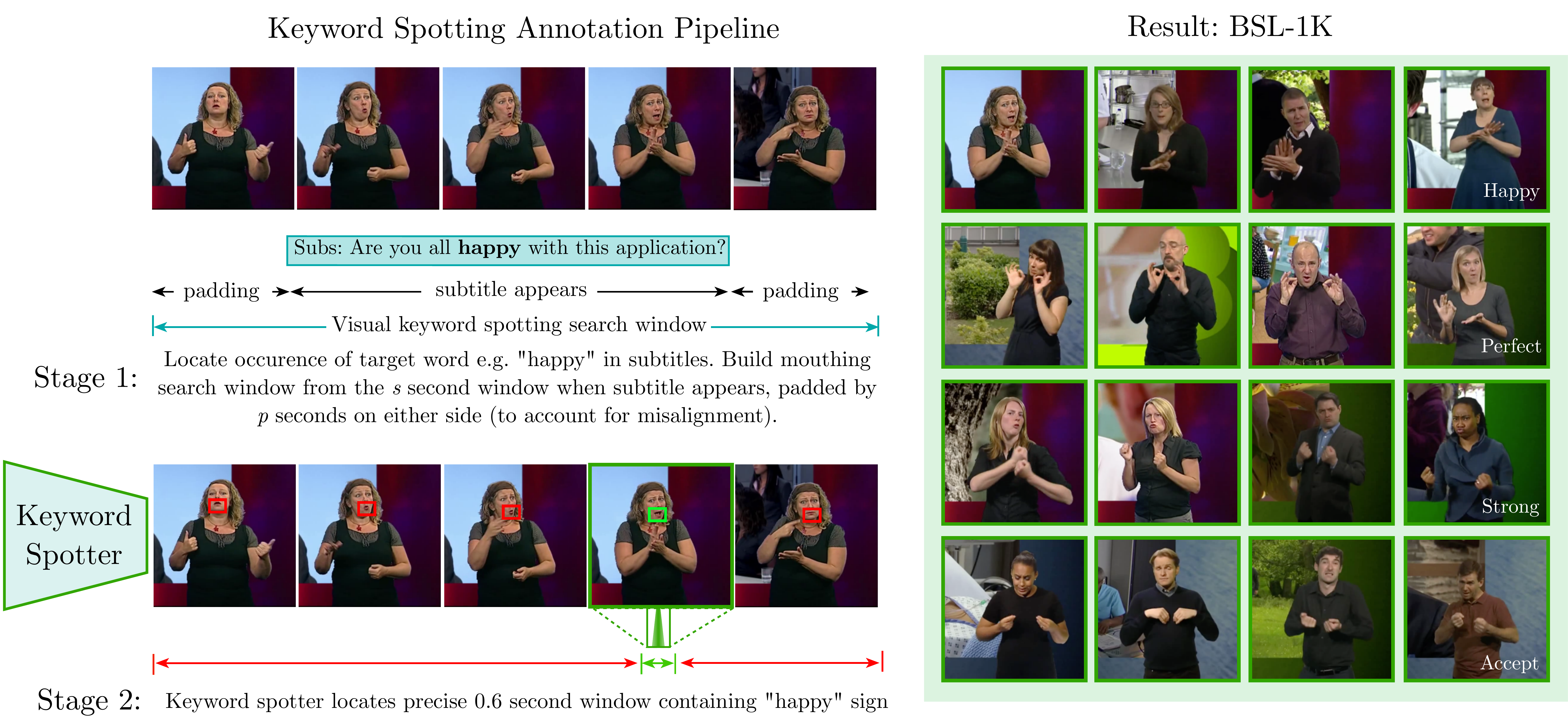}
    \caption{\textbf{Mouthing-based sign annotation from~\cite{Albanie20}:} (Left, the annotation pipeline): Stage 1:
            for a given sign (e.g.~“happy”), each instance of the word in the subtitles provides a candidate temporal segment where the sign may occur (the subtitle timestamps are padded by several seconds to account for the %
            asynchrony %
            between audio-aligned subtitles and signing interpretation);
            Stage 2: a mouthing visual keyword spotter uses the lip movements of the signer to perform precise localisation of the sign within this window. (Right): Examples of localised signs through mouthings from the BSL-1K dataset—
            produced by applying keyword spotting for a vocabulary of 1K words.}
    \label{fig:kws}
\end{figure*}

Our objective, illustrated in Fig.~\ref{fig:kws-task}, is to discover all occurrences of a 
given keyword in a collection of continuous signing sequences.  To do so, we assume access 
to: (i) a subtitled collection of videos containing continuous signing, $\mathcal{S} = \{(x_i, s_i ) : i \in \{1, \dots, I\}, x_i \in \mathcal{X}_{\mathfrak{L}}, s_i \in \mathcal{T}\}$; (ii) 
a sparse collection of temporal sub-segments of these videos that have been annotated with 
their corresponding word, $\mathcal{M} = \{(x_k, v_k) : k \in \{1, \dots, K\}, v_k \in \mathcal{V}_\mathfrak{L}, x_k \in \mathcal{X}_\mathfrak{L}^{\text{single}}, \exists (x_i, s_i) \in \mathcal{S} \, s.t. \, x_k \subseteq x_i \}$; 
(iii) a curated \textit{dictionary} of signing instances $\mathcal{D} = \{(x_j, v_j) : j \in \{1, \dots, J\}, x_j \in \mathcal{X}_{\mathfrak{L}}^{\text{single}}, v_j \in \mathcal{V}_\mathfrak{L}\}$.
To address the sign spotting task, we propose to learn a \textit{data representation} $f: \mathcal{X}_\mathfrak{L} \rightarrow \mathbb{R}^d$ that maps video segments to vectors such 
that they are \textit{discriminative} for sign spotting and \textit{invariant} to other 
factors of variation.  Formally, for any labelled pair of video segments $(x, v), (x', v')$ 
with $x, x' \in \mathcal{X}_\mathfrak{L}$ and $v, v' \in \mathcal{V}_\mathfrak{L}$, we seek 
a data representation, $f$, that satisfies the constraint  $\delta_{f(x) f(x')} = \delta_{v v'}$,
where $\delta$ represents the Kronecker delta. 

\subsection{Sparse annotations from mouthing cues} %
\label{subsection:visual-kws}

As the source of temporal video segments with corresponding word annotations, $\mathcal{M}$,
we make use of automatic annotations that were collected as part of our prior work on 
visual keyword spotting with mouthing cues~\cite{Albanie20}, which we briefly recap here. 
Signers 
sometimes mouth a word while simultaneously signing it, as an additional signal~\cite{Bank2011,sutton-spence_woll_1999,Sutton-Spence2007}, performing similar lip patterns as for the spoken word.
Fig.~\ref{fig:kws} presents an overview of how we use such mouthings to spot signs.

As a starting point for this approach, we assume access to TV footage that is accompanied 
by: (i) a frontal facing sign language interpreter, who provides a translation of the 
spoken content of the video, and (ii) a subtitle track, representing a direct transcription 
of the spoken content.
The method of~\cite{Albanie20} first searches among the subtitles for any occurrences of 
``keywords'' from a given vocabulary.
Subtitles containing these keywords provide a set of candidate temporal windows in which 
the interpreter may have produced the sign corresponding to the keyword (see Fig.~\ref{fig:kws}, Left, Stage 1). However, these 
temporal windows are difficult to make use of directly since: 
(1)~the occurrence of a keyword in a subtitle does not ensure the presence of the 
corresponding sign in the signing sequence,
(2)~the subtitles themselves are not precisely aligned with the signing, and can differ in 
time by several seconds.
To address these issues,~\cite{Albanie20} demonstrated that the sign corresponding to a 
particular keyword can be localised within a candidate temporal window -- given by the \textit{padded} subtitle timings to account for the asynchrony %
between the audio-aligned subtitles and signing interpretation -- by searching for its \textit{spoken components}~\cite{sutton-spence_woll_1999} amongst the mouth movements of 
the interpreter.
While there are challenges associated with using spoken components as a cue (signers do not 
typically mouth continuously and may only produce mouthing patterns that correspond to a 
portion of the keyword~\cite{sutton-spence_woll_1999}), it has the significant advantage of 
transforming the general annotation problem from classification (i.e., ``which sign is 
this?'') into the much easier problem of localisation (i.e., ``find a given token amongst a 
short sequence'').
In~\cite{Albanie20}, the visual keyword spotter uses the candidate temporal window with the target keyword to estimate the probability that the sign was mouthed at each time step. If the peak probability over time is above a threshold parameter, the predicted location of the sign is taken as the 0.6 second window starting before the position of the peak probability (see Fig.~\ref{fig:kws}, Left, Stage 2). For building the BSL-1K dataset, \cite{Albanie20} uses a probability threshold of 0.5 and runs the visual keyword spotter with a 
vocabulary of 1,350 keywords across 1,000 hours of signing.
A further filtering step is performed on the vocabulary to ensure that each word included in the
dataset is represented with high confidence (at least one instance with confidence
0.8) in the training partition, which produces a final dataset vocabulary of 1,064
words. The resulting BSL-1K dataset has 273K mouthing annotations, some of which are illustrated in Fig.~\ref{fig:kws} (right).
We 
employ these annotations directly to form the set $\mathcal{M}$ in this work.

\subsection{Integrating cues through multiple instance learning }
\label{subsection:mil}

\begin{figure*}[t]
  \centering
  \includegraphics[width=0.98\textwidth]{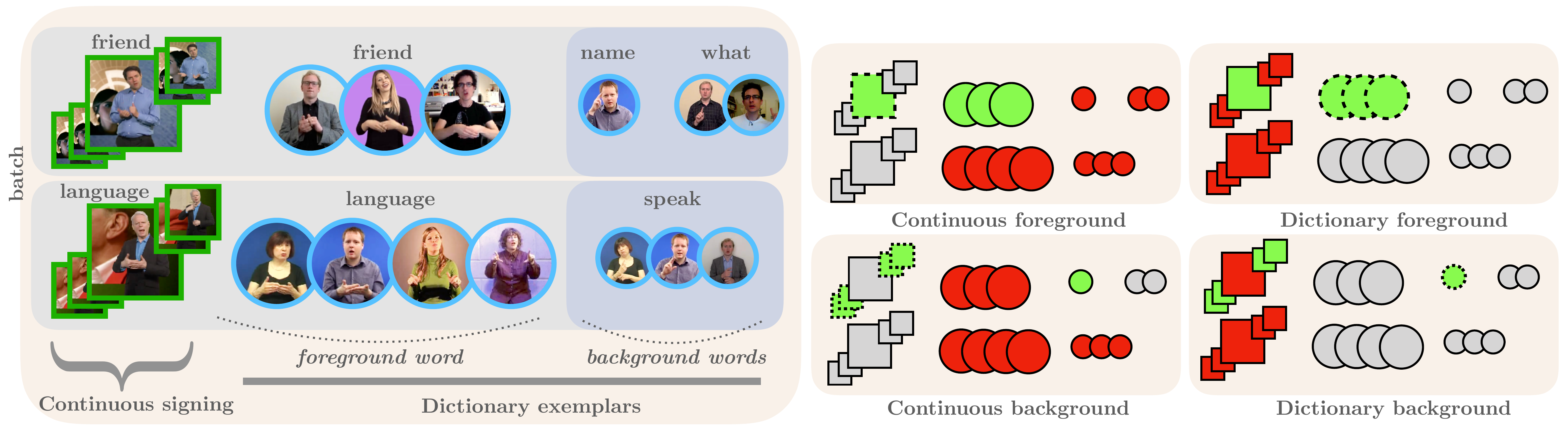}
  \caption{\textbf{Batch sampling and positive/negative pairs:} We illustrate the formation 
  of a batch when jointly training on continuous signing video (squares) and dictionaries 
  of isolated signing (circles).
  \textbf{Left:} For each continuous video, we sample the dictionaries corresponding to the 
  labelled word (foreground), as well as %
  to the rest of the subtitles (background). 
  \textbf{Right:} We construct positive/negative pairs by anchoring at 4 different portions 
  of a batch item: continuous foreground/background and dictionary foreground/background. 
  Positives and negatives (defined across continuous and dictionary domains)
  are green and red, respectively; anchors have a dashed border (see
  \if\sepappendix1{Appendix~C.2}
  \else{Appendix~\ref{app:subsec:math}}
  \fi
  for details).
  }
  \label{fig:sampling2}
\end{figure*}

To learn $f$, we must address several challenges.  First, as noted in Sec.~\ref{sec:intro}, 
there may be a considerable distribution shift between the dictionary videos of isolated 
signs in $\mathcal{D}$ and the co-articulated signing videos in $\mathcal{S}$.  Second, 
sign languages often contain multiple sign variants for a single written word (e.g., resulting 
from regional variations and synonyms).  Third, since the subtitles in $\mathcal{S}$ are only 
weakly aligned with the sign sequence, we must learn to associate signs and words from a 
noisy signal that lacks temporal localisation.  Fourth, the localised annotations provided 
by $\mathcal{M}$ are sparse, and therefore we must make good use of the remaining segments 
of subtitled videos in $\mathcal{S}$ if we are to learn an effective representation.

Given full supervision, we could simply adopt a pairwise metric learning approach to align 
segments from the videos in $\mathcal{S}$ with dictionary videos from $\mathcal{D}$ by 
requiring that $f$ maps a pair of isolated and co-articulated signing segments to the same 
point in the embedding space if they correspond to the same sign (\textit{positive} pairs) 
and apart if they do not (\textit{negative} pairs).
As noted above, in practice we do not have access to positive pairs because: (1) for any 
annotated segment $(x_k, v_k) \in \mathcal{M}$, we have a set of potential sign variations 
represented in the dictionary (annotated with the common label $v_k$), rather than a single 
unique sign; (2) since $\mathcal{S}$ provides only weak supervision, even when a word is 
mentioned in the subtitles we do not know where it appears in the continuous signing 
sequence (if it appears at all). 
These ambiguities motivate a Multiple Instance Learning~\cite{dietterich1997solving} (MIL) 
objective. Rather than forming positive and negative pairs, we instead form positive 
\textit{bags} of pairs, $\mathcal{P}^{\text{bags}}$, in which we expect at least one 
pairing between a segment from a video in $\mathcal{S}$ and a dictionary video from 
$\mathcal{D}$ to contain the same sign, and negative bags of pairs, 
$\mathcal{N}^{\text{bags}}$, in which we expect no (video segment, dictionary video) pair 
to contain the same sign.  To incorporate the available sources of supervision into this 
formulation, we consider two categories of positive and negative bag formations, described 
next (%
a formal mathematical description of the positive and 
negative bags described below is deferred to
\if\sepappendix1{Appendix~C.2).}
\else{Appendix~\ref{app:subsec:math}).}
\fi

\noindent \textbf{Watch and Lookup: using sparse annotations and dictionaries}. Here, we 
describe a baseline where we assume no subtitles are available. To learn $f$ from 
$\mathcal{M}$ and $\mathcal{D}$, we define each positive bag as the set of possible pairs 
between a \textit{labelled} (\textit{foreground}) temporal segment of a continuous video 
from $\mathcal{M}$ and the examples of the corresponding sign in the dictionary (green 
regions in
\if\sepappendix1{Fig~A.2).}
\else{Fig~\ref{app:fig:sampling_milnce}).}
\fi
The key assumption here is that each labelled sign segment  from $\mathcal{M}$ matches 
\textit{at least one} sign variation in the dictionary.
Negative bags are constructed by (i) anchoring on a continuous
foreground segment and selecting dictionary examples
corresponding to different words
from other batch items; (ii) anchoring on
a dictionary foreground set and
selecting continuous foreground segments from other batch items
(red regions in
\if\sepappendix1{Fig~A.2).}
\else{Fig~\ref{app:fig:sampling_milnce}).}
\fi
To maximize the number of negatives
within one minibatch, we sample a different word per batch item.

\noindent \textbf{Watch, Read and Lookup: using sparse annotations, subtitles and dictionaries}. 
Using just the labelled sign segments from $\mathcal{M}$ to construct bags has a 
significant limitation: $f$ is not encouraged to represent signs beyond the initial 
vocabulary represented in $\mathcal{M}$.  We therefore look at the subtitles (which contain 
words beyond $\mathcal{M}$) to construct additional bags. We determine more positive bags 
between the set of \textit{unlabelled (background)} segments in the continuous footage and 
the set of dictionaries corresponding to the background words in the subtitle (green 
regions in Fig.~\ref{fig:sampling2},  right-bottom). Negatives (red regions in 
Fig.~\ref{fig:sampling2}) are formed as the complements to these sets by (i) pairing 
continuous background segments with dictionary samples that can be excluded as matches 
(through subtitles) and (ii) pairing background dictionary entries with the foreground 
continuous segment.
In both cases, we also define negatives from other batch items
by selecting pairs where the word(s) have no overlap, e.g.,
in~Fig.~\ref{fig:sampling2}, the dictionary examples for the background word `speak' from 
the second batch item are negatives
for the background continuous segments from the first batch item, corresponding to the 
unlabelled words `name' and `what' in the subtitle.

To assess the similarity of two embedded video segments, we employ a similarity function 
$\psi: \mathbb{R}^d \times \mathbb{R}^d \rightarrow \mathbb{R}$ whose value increases as 
its arguments become more similar (in this work, we use cosine similarity). For notational 
convenience below, we write $\psi_{ij}$ as shorthand for $\psi(f(x_i), f(x_j))$. 
To learn $f$, we consider a generalization of the InfoNCE 
loss~\cite{oord2018representation,wu2018unsupervised} (a non-parametric softmax loss 
formulation of Noise Contrastive Estimation~\cite{gutmann2010noise}) recently proposed 
by~\cite{miech2020end} as MIL-NCE loss:

\begin{align}
    \mathcal{L} = - \mathbb{E}_i \Bigg[ \log \frac{\sum\limits_{(j,k) \in \mathcal{P}(i)} e^{\psi_{jk}/ \tau}}{\sum\limits_{(j,k) \in \mathcal{P}(i)} e^{\psi_{jk}/ \tau} +  \sum\limits_{(l,m) \in \mathcal{N}(i)} e^{\psi_{lm}/ \tau}} \Bigg], \label{eqn:mil-nce}
\end{align}

where $\mathcal{P}(i) \in \mathcal{P}^{\text{bags}}$, $\mathcal{N}(i) \in \mathcal{N}^{\text{bags}}$,
$\tau$, often referred to as the \textit{temperature}, is set as a hyperparameter (we 
explore the effect of its value in Sec.~\ref{sec:experiments}).

\subsection{Implementation details }\label{subsection:implementation}

In this section, we provide details for the learning framework covering the embedding 
architecture, sampling protocol and optimization procedure.

\noindent{\textbf{Embedding architecture.}} The architecture comprises an I3D 
spatio-temporal trunk network~\cite{Carreira2017} to which we attach an MLP consisting of 
three linear layers separated by leaky ReLU activations (with negative slope 0.2) and a 
skip connection. The trunk network takes as input 16 frames from a $224\times224$ 
resolution video clip and produces $1024$-dimensional embeddings which are then projected 
to 256-dimensional sign spotting embeddings by the MLP.
More details about the embedding architecture can be found in
\if\sepappendix1{Appendix~C.1.}
\else{Appendix~\ref{app:subsec:arch}.}
\fi

\noindent{\textbf{Joint pretraining.}} The I3D trunk parameters are initialised by 
pretraining for sign classification jointly over the sparse annotations $\mathcal{M}$ of a 
continuous signing dataset (\bslonek{}~\cite{Albanie20}) and examples from a sign 
dictionary dataset (\bsldict) which fall within their common vocabulary.
Since we find that dictionary videos of isolated signs tend to be performed more slowly, we 
uniformly sample 16 frames from each dictionary video with a random shift and random frame 
rate $n$ times, where $n$ is proportional to the length of the video, and pass these clips 
through the I3D trunk then average the resulting vectors before they are processed by the 
MLP to produce the final dictionary embeddings. We find that this form of random sampling 
performs better than sampling 16 consecutive frames from the isolated signing videos (see
\if\sepappendix1{Appendix~C.1}
\else{Appendix~\ref{app:subsec:arch}}
\fi
for more details).
During pretraining, minibatches of size 4 are used; and colour, scale and horizontal flip 
augmentations are applied to the input video, following the procedure described 
in~\cite{Albanie20}.
The trunk parameters are then frozen and the MLP outputs are used as embeddings. Both 
datasets are described in detail in Sec.~\ref{subsection:datasets}.

\noindent{\textbf{Minibatch sampling.}} To train the MLP given the pretrained I3D features, 
we sample data by first iterating over the set of labelled segments comprising the sparse 
annotations, $\mathcal{M}$, that accompany the dataset of continuous, subtitled sampling to 
form minibatches.
For each continuous video, we sample 16 consecutive frames around the annotated timestamp 
(more precisely a random offset within 20 frames before, 5 frames after, following the 
timing study in~\cite{Albanie20}). We randomly sample 10 additional 16-frame clips from 
this video outside of the labelled window, i.e., continuous background segments.
For each subtitled sequence, we sample the dictionary entries for all subtitle words that 
appear in $\mathcal{V}_{\mathfrak{L}}$ (see Fig.~\ref{fig:sampling2} for a sample batch 
formation).

Our minibatch comprises 128 sequences of continuous signing and their corresponding 
dictionary entries (we investigate the impact of batch size in 
Sec.~\ref{subsection:ablations}). The embeddings are then trained by minimising the loss 
defined in Eqn.(\ref{eqn:mil-nce}) in conjunction with 
positive bags, $\mathcal{P}_{\text{}}^{\text{bags}}$, and negative bags, 
$\mathcal{N}_{\text{}}^{\text{bags}}$, which are constructed on-the-fly for each minibatch 
(see Fig.~\ref{fig:sampling2}).

\noindent{\textbf{Optimization.}} We use a SGD optimizer with an initial learning rate of 
$10^{-2}$ to train the embedding architecture. The learning rate is decayed twice by a 
factor of 10 (at epochs 40 and 45). We train all models, including baselines and ablation 
studies, for 50 epochs at which point we find that learning has always converged.

\noindent{\textbf{Test time.}} To perform spotting, we obtain the embeddings learned with the MLP.
For the dictionary, we have a single embedding averaged over the video.
Continuous video embeddings are obtained with sliding window (stride 1) on the entire 
sequence. We show the importance of using such a dense stride for a precise localisation
in our ablations (Sec.~\ref{subsection:ablations}).
However, for simplicity, all qualitative visualisations are performed with continuous 
video embeddings obtained with a sliding window of stride 8.

We calculate the cosine similarity score between the continuous signing sequence embeddings 
and the embedding for a given dictionary video. We determine the location with the maximum 
similarity as the location of the queried sign. We maintain embedding sets of all variants 
of dictionary videos for a given word and choose the best match as the one with the highest 
similarity.

\section{Experiments} \label{sec:experiments}

\begin{table*}[t]
 \centering
\begin{tabularx}{0.99\linewidth}{c l r r r c}
\toprule
Dataset & Split & \#Videos & Vocabulary & \#Signers & Examples\\ 
\midrule
 \multirow{7}{*}{\raisebox{0.1cm}{\bslonek\cite{Albanie20}}}
 & \multirow{1}{*}{\trainrec}
 & \multirow{1}{*}{169K}
 & \multirow{1}{*}{1,064} 
 & \multirow{1}{*}{36}
 & \multirow{5}{*}{\raisebox{-1.5cm}{\includegraphics[scale=0.45]{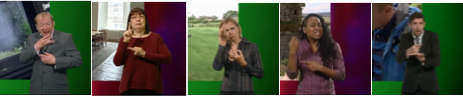}}}
 \\[0.05cm]
 & \multirow{1}{*}{\testrec}
 & \multirow{1}{*}{2,103}
 & \multirow{1}{*}{334}
 & \multirow{1}{*}{4}
 \\[0.05cm]
 & \multirow{1}{*}{\testrecnew}
 & \multirow{1}{*}{36,854}
 & \multirow{1}{*}{950}
 & \multirow{1}{*}{4}
 \\[0.05cm]
 \cline{2-5}\\[-0.1cm]
 & \multirow{1}{*}{\trainspot}
 & \multirow{1}{*}{78,211}
 & \multirow{1}{*}{1,064}
 & \multirow{1}{*}{36}
 \\[0.05cm]
 & \multirow{1}{*}{\testspot}
 & \multirow{1}{*}{1,834} %
 & \multirow{1}{*}{264}
 & \multirow{1}{*}{4}
 \\[0.05cm]
\midrule
\multirow{5}{*}{\bsldict}
& & & & &
\multirow{4}{*}{\raisebox{0.1cm}{\includegraphics[scale=0.45]{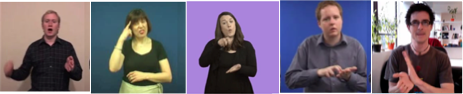}}}
\\[0.05cm]
 & \multirow{2}{*}{Full}
 & \multirow{2}{*}{14,210}
 & \multirow{2}{*}{9,283}
 & \multirow{2}{*}{124} %
 & 
 \\[0.05cm]
 & \multirow{2}{*}{1K-subset}
 & \multirow{2}{*}{2,963}
 & \multirow{2}{*}{1,064}
 & \multirow{2}{*}{70}
 \\[0.05cm]
 \\
\bottomrule
\end{tabularx}
    \caption{
         \textbf{Datasets:} We provide (i) the number of individual signing videos,
         (ii) the vocabulary size of the \textit{annotated} %
         signs, and (iii) the number of signers
         for several subsets of \bslonek{} and \bsldict. \bslonek{}
         is large in the number of annotated signs whereas \bsldict{} is large in the vocabulary size.
         Note that \bslonek{} is constructed differently depending on
         whether it is used for the task of recognition or retrieval: for retrieval, longer signing
         sequences are used around individual localised signs as described
         in Sec.~\ref{subsection:datasets}.
     }
    \label{tab:datasets}
\end{table*}

In this section, we first present the datasets used in this work (including the contributed 
\bsldict{} dataset) in Sec.~\ref{subsection:datasets}, followed by the evaluation protocol 
in Sec.~\ref{subsec:evaluation}.
We then illustrate the benefits of the \textit{Watch, Read and Lookup} learning framework 
for sign spotting
against several baselines (Sec.~\ref{subsection:comparison})
with a comprehensive ablation study that validates our design choices 
(Sec.~\ref{subsection:ablations}).
Next, we investigate three applications of our method in 
Sec.~\ref{subsection:applications}, showing that it can be used to (i)~not only spot signs, 
but also identify the specific sign variant that was used, (ii)~label 
sign instances in continuous signing footage given the associated subtitles, and 
(iii)~discover \say{faux amis} between different sign languages. 
We then provide experiments on sign language recognition, %
    significantly improving the state of the art by applying our 
    labelling technique to obtain more training examples automatically
    (Sec.~\ref{subsection:recognition} and Sec.~\ref{subsection:bobsl}).
    Finally, we discuss limitations of our sign spotting technique using dictionaries
    (Sec.~\ref{subsection:limitations}).

\subsection{Datasets}\label{subsection:datasets}
Although our method is conceptually applicable to a number of sign languages, in this work 
we focus primarily on BSL, the sign language of British deaf communities. We
use \bslonek~\cite{Albanie20}, a large-scale, subtitled and sparsely annotated dataset of 
more than 1,000 hours of continuous signing which offers an ideal setting in which to 
evaluate the effectiveness of the \textit{Watch, Read and Lookup} sign spotting framework.  
To provide dictionary data for the \textit{lookup} component of our approach, we also 
contribute~\bsldict{}, a diverse visual
dictionary of signs. These two datasets are summarised in Tab.~\ref{tab:datasets} and 
described in more detail below.
We further include experiments on a new dataset, BOBSL~\cite{Albanie2021bobsl},
    which we describe in Sec.~\ref{subsection:bobsl} together with results. The
    BOBSL dataset has similar properties to \bslonek{}.

\noindent\textbf{\bslonek}~\cite{Albanie20} comprises over 1,000 hours of video of continuous sign-language-interpreted %
television broadcasts, with accompanying subtitles of the audio content. In~\cite{Albanie20}, this data is processed for the task of individual sign recognition: a visual keyword spotter is applied to signer mouthings %
giving a total of 273K sparsely localised sign annotations
from a vocabulary of 1,064 signs (169K in the training partition as shown in Tab.~\ref{tab:datasets}). 
Please refer to Sec.~\ref{subsection:visual-kws} and~\cite{Albanie20} for more 
details on the automatic annotation pipeline. We refer to 
Sec.~\ref{subsection:recognition} for a description of the BSL-1K sign recognition 
benchmark (\testrec{} and \testrecnew{} in Tab.~\ref{tab:datasets}).

In this work,
we process this data for the task of retrieval, extracting long videos with associated subtitles. In particular, we pad $\pm$2 
seconds around the subtitle timestamps and we add the corresponding video to our training 
set if there is a sparse annotation from mouthing falling within this time window -- we assume this constraint indicates that the 
signing is reasonably well-aligned with its subtitles. We further consider 
only the videos whose subtitle duration is longer than 2 seconds. For testing, we use the 
automatic test set (corresponding to mouthing locations with confidences above 0.9). Thus 
we obtain 78K training (\trainspot) and 2K %
test (\testspot) videos as shown in Tab.~\ref{tab:datasets}, each of which has a subtitle of 8 words on 
average and 1 sparse mouthing annotation. \\

\noindent\textbf{\bsldict}. BSL dictionary videos are collected from a BSL sign aggregation 
platform \url{signbsl.com}~\cite{signbslcom}, giving us a total of 14,210 video clips for a 
vocabulary of 9,283 signs. Each sign is typically performed several times by different 
signers, often in different ways. The dictionary videos are linked %
from 28 known 
website sources and each source has at least 1 signer. We used face embeddings computed 
with SENet-50~\cite{hu2019squeeze} (trained on VGGFace2~\cite{Cao18}) to cluster signer 
identities and manually verified that there are a total of 124 %
different signers. The 
dictionary videos are of isolated signs (as opposed to co-articulated in \bslonek): this 
means (i) the start and end of the video clips usually consist of a still signer pausing, 
and (ii) the sign is performed at a much slower rate for clarity. We first trim the sign 
dictionary videos, using body keypoints estimated with OpenPose~\cite{cao2018openpose} 
which indicate the start and end of wrist motion, to discard frames where the signer is 
still. With this process, the average number of frames per video drops from 78 to 56 (still 
significantly larger than co-articulated signs).
To the best of our knowledge, \bsldict{} is the first curated, BSL sign dictionary dataset 
for computer vision research. A collection of metadata associated for the
\bsldict{} dataset is made publicly available, as well as
our pre-computed video embeddings from this work.

For the experiments in which 
\bsldict{} is filtered to the 1,064 vocabulary of \bslonek{}, we have
3K
videos as shown in Tab.~\ref{tab:datasets}.
Within this subset, each sign has between 1 and 10 examples (average of 3).

\subsection{Evaluation protocols}\label{subsec:evaluation}
\noindent{\textbf{Protocols.}} We define two settings: (i) training with the entire 1,064 
vocabulary of annotations in \bslonek{}; and (ii) training on a subset with 800 signs. The 
latter is needed to assess the performance on novel signs, for which we do not have access 
to co-articulated labels at training. We thus use the remaining 264 words for testing. This 
test set is therefore common to both training settings, it is either `seen' or `unseen' at 
training.
However, we do not limit the vocabulary of the dictionary as a practical assumption,
for which we show benefits.

\noindent{\textbf{Metrics.}}
The performance is evaluated based on ranking metrics as in retrieval.
For every sign $s_i$ in the test vocabulary, we first select the \bslonek{} test set clips 
which have a mouthing annotation of $s_i$ and then record the percentage of times that a dictionary 
clip of $s_i$ appears in the first 5 retrieved results, this is the ‘Recall at 5’ 
(R@5). This is motivated by the fact that different English words can correspond to the 
same sign, and vice versa. We also report mean average precision (mAP). For each video 
pair, the match is considered correct if (i) the dictionary clip corresponds to $s_i$  and 
the \bslonek{} video clip has a mouthing annotation of $s_i$, and (ii) if the predicted 
location of the sign in the \bslonek{} video clip, i.e., the time frame where the maximum 
similarity occurs, lies within certain frames around the ground truth mouthing timing. In 
particular, we determine the correct interval to be defined between 20 frames before and 5 
frames after the labelled time (based on the study in~\cite{Albanie20}).
Finally, because the \bslonek{} test set is class-unbalanced, we report performances averaged over 
the test classes.

\begin{table*}[t]
    \setlength{\tabcolsep}{8pt}
    \centering
    \resizebox{0.99\linewidth}{!}{
        \begin{tabular}{llll|ll}
            \toprule
            & & \multicolumn{2}{c|}{Train (1064)} & \multicolumn{2}{c}{Train (800)} \\
            \midrule
            & & \multicolumn{2}{c|}{Seen (264)} & \multicolumn{2}{c}{Unseen (264)} \\
            Embedding arch. & Supervision & mAP & R@5 & mAP & R@5 \\
            \midrule
            \baselineDict{} & Classification & \phantom{0}2.68 & \phantom{0}3.57 & \phantom{0}1.21 & \phantom{0}1.29 \\
            \baselineBbc{} \cite{Albanie20} & Classification & 13.09 & 17.25 & \phantom{0}6.74 & \phantom{0}8.94 \\
            \baselineBbcDict{} & Classification & 19.81 &  25.57 & \phantom{0}4.81 & \phantom{0}6.89 \\

            \baselineBbcDict{}+MLP & Classification & 37.13 $\pm$ 0.29 & 39.68 $\pm$ 0.57 & 10.33 $\pm$ 0.43 & 13.33 $\pm$ 1.11 \\
            \midrule
            \baselineBbcDict{}+MLP & InfoNCE & 43.59 $\pm$ 0.76 & 52.59 $\pm$ 0.75 & 11.40 $\pm$ 0.42 & 14.76 $\pm$ 0.40 \\
            \baselineBbcDict{}+MLP & Watch-Lookup & 44.72 $\pm$ 0.85 & 55.51 $\pm$ 2.17 & 11.02 $\pm$ 0.27 & 15.03 $\pm$ 0.45 \\
            \baselineBbcDict{}+MLP & Watch-Read-Lookup & \textbf{47.93} $\pm$ 0.20 & \textbf{60.76} $\pm$ 1.45 & \textbf{14.86} $\pm$ 1.29 & \textbf{19.85} $\pm$ 1.94 \\
            \bottomrule
        \end{tabular}
    }
    \caption{\textbf{The effect of the loss formulation:} Embeddings learned with the 
        classification loss are suboptimal since they are not trained for matching the two 
        domains. Contrastive-based loss formulations (NCE) significantly improve, 
        particularly when we adopt the multiple-instance variant introduced as our 
        Watch-Read-Lookup framework of multiple supervisory signals.
        We report the relatively cheaper MLP-based models with three random seeds for each model and report the mean and the standard deviation.
    }
    \label{tab:loss}
\end{table*}

\subsection{Comparison to baselines}\label{subsection:comparison}
In this section, we evaluate different components of our approach.
We first compare our contrastive learning approach with classification baselines.
Then, we investigate the effect of our multiple-instance loss formulation.
Finally, we report performance on a sign spotting benchmark.

\noindent\textbf{I3D baselines.}
We start by evaluating baseline I3D models trained with classification
on the task of spotting, using the embeddings before the classification layer.
We have three variants in Tab.~\ref{tab:loss}: (i) \baselineBbc{} provided by \cite{Albanie20}
which is trained only on the \bslonek{} dataset, and we also train (ii) \baselineDict{}
and (iii) \baselineBbcDict{}. Training only on \bsldict{} (\baselineDict{})
performs significantly worse due to the few examples available per class and the domain gap that must be bridged to spot co-articulated signs, suggesting that dictionary samples alone do not suffice to solve the task. We observe
improvements with fine-tuning
\baselineBbc{} jointly on the two datasets (\baselineBbcDict{}),
which becomes our base feature extractor for the remaining experiments
to train a shallow MLP.

\noindent\textbf{Loss formulation.}
We first train the MLP parameters on top of the frozen I3D trunk with classification to establish a baseline in a comparable setup.
Note that, this shallow architecture can be trained with larger batches than I3D.
Next, we investigate variants of our loss to learn a joint sign embedding between 
\bslonek{} and \bsldict{} video domains: (i) standard single-instance 
InfoNCE~\cite{oord2018representation,wu2018unsupervised} loss which pairs each \bslonek{} 
video clip with \textit{one} positive \bsldict{} clip of the same sign, (ii) Watch-Lookup 
which considers
multiple positive dictionary candidates, but does not consider subtitles (therefore limited 
to the annotated video clips). Tab.~\ref{tab:loss}
summarises the results. Our Watch-Read-Lookup formulation which
effectively combines multiple sources of supervision
in a multiple-instance framework outperforms the other baselines
in both \textit{seen} and \textit{unseen} protocols.

\noindent\textbf{Extending the vocabulary.} The results presented so far
were using the same vocabulary for both continuous and dictionary datasets. In reality,
one can assume access to the entire vocabulary in the dictionary, but obtaining
annotations for the continuous videos is prohibitive.
Tab.~\ref{tab:vocab}
investigates removing the vocabulary limit on the dictionary side, but keeping
the continuous annotations vocabulary at 800 signs. We show that using the full 9k 
vocabulary from
\bsldict{} %
improves the results on the unseen setting.

\noindent\textbf{\bslonek{} sign spotting benchmark.} Although our learning framework 
primarily targets good performance on unseen continuous signs, it can also be naively 
applied to the (closed-vocabulary) sign spotting benchmark proposed by~\cite{Albanie20}.
The sign spotting benchmark requires a model to localise every instance belonging to a given set of sign classes (334 in total) within long sequences of untrimmed footage.
The benchmark is challenging because each sign appears infrequently (corresponding to approximately one positive instance in every 90 minutes of continuous signing).
We 
evaluate the performance of our Watch-Read-Lookup model and achieve a score of 0.170 mAP, 
outperforming the previous state-of-the-art performance of 0.160 mAP~\cite{Albanie20}.

\begin{table}[t]
    \setlength{\tabcolsep}{8pt}
    \centering
    \resizebox{0.99\linewidth}{!}{
        \begin{tabular}{llcc}
            \toprule
            Supervision & Dictionary Vocab & mAP & R@5 \\
            \midrule
            
            Watch-Read-Lookup & 800 training vocab & 14.86 $\pm$ 1.29 & 19.85 $\pm$ 1.94 \\
            Watch-Read-Lookup & 9k full vocab & \textbf{15.82} $\pm$ 0.48 & \textbf{21.67} $\pm$ 0.72 \\
            \bottomrule
        \end{tabular}
    }
    \caption{\textbf{Extending the dictionary vocabulary:} We show the benefits of sampling 
        dictionary videos outside of the sparse annotations, using subtitles. Extending the 
        lookup to the dictionary from the subtitles to the full vocabulary of \bsldict{} 
        brings significant improvements for novel signs (the training uses sparse 
        annotations for the 800 words, and the remaining 264 for test).
    }
    \label{tab:vocab}
\end{table}

\setlength{\tabcolsep}{1pt}
\begin{figure*}[t]
    \centering
    \begin{tabular}{llll}
    (a) & \raisebox{-.5\height}{\includegraphics[width=0.45\textwidth]{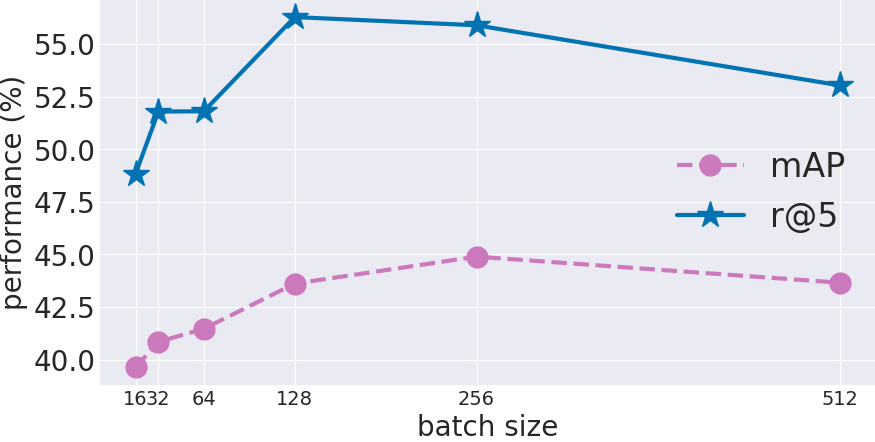}} & (b) & \raisebox{-.5\height}{\includegraphics[width=0.45\textwidth]{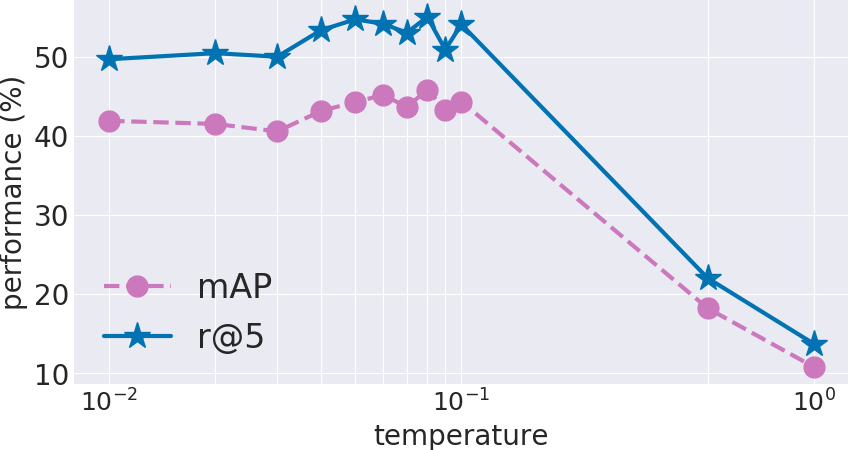}} %
    \end{tabular}
    \caption{The effect of (a) the \textbf{batch size} that determines the number of 
        negatives across sign classes and (b) the \textbf{temperature} hyper-parameter for 
        the MIL-NCE loss in Watch-Lookup against mAP and R@5 (trained on the full 1064 
        vocab.)}
    \label{fig:hyperparams}
\end{figure*}

\subsection{Ablation study}\label{subsection:ablations}
We provide ablations for the learning hyperparameters, such as the batch size and the 
temperature; the mouthing confidence threshold as the training data selection parameter;
and the stride parameter of the sliding window at test time.

\noindent\textbf{Batch size.} Next, we investigate the effect of increasing the number of 
negative pairs by increasing the batch size when training with Watch-Lookup on 1,064 
categories. We observe in Fig.~\ref{fig:hyperparams}(a) an improvement in performance 
with a greater number of negatives before saturating.
Our final Watch-Read-Lookup model has high memory requirements, for which we use 128 batch 
size. Note that the effective size of the batch with our sampling is larger
due to sampling extra video clips corresponding to subtitles.

\noindent\textbf{Temperature.} Finally, we analyze the impact of the temperature 
hyperparameter $\tau$ on the performance of Watch-Lookup.
We conclude from Fig.~\ref{fig:hyperparams}(b) that setting $\tau$ to values between
[0.04 - 0.10] does not impact the performance significantly;
therefore,
we keep $\tau = 0.07$ following the previous work
\cite{wu2018unsupervised,he2020momentum} for all other experiments. 
    However, values outside this range negatively impact the performance,
    especially for high values, i.e., $\{0.50, 1.00\}$;
we observe a major decrease in 
performance when $\tau$ approaches 1.

\noindent\textbf{Mouthing confidence threshold at training.} As explained in 
Sec.~\ref{subsection:visual-kws},
the sparse annotations from the \bslonek{} dataset are obtained
automatically by running a visual keyword spotting method based on mouthing cues.
The dataset provides a confidence value associated with
each label ranging between $0.5$ and $1.0$. Similar to \cite{Albanie20},
we experiment with different thresholds to determine the training set.
Lower thresholds result in a noisier but larger training set.
From Tab.~\ref{tab:mouthingthres}, we conclude that
$0.5$ mouthing confidence threshold performs the best.
This is in accordance with the conclusion from \cite{Albanie20}.

\begin{table}
    \setlength{\tabcolsep}{8pt}
    \centering
    \resizebox{0.99\linewidth}{!}{
        \begin{tabular}{llcc}
            \toprule
            Mouthing confidence & Training size & mAP & R@5 \\
            \midrule
            0.9 & 10K & 37.55 & 47.54 \\ %
            0.8 & 21K & 39.49 & 48.84 \\ %
            0.7 & 33K & 41.87 & 51.15 \\ %
            0.6 & 49K & 42.44 & 52.42 \\ %
            0.5 & 78K & \textbf{43.65} & \textbf{53.03} \\ %
            \bottomrule
        \end{tabular}
    }
    \caption{\textbf{Mouthing confidence threshold:} The results suggest that lower 
            confidence automatic annotations of \bslonek{} provide better training, by 
            increasing the amount of data (training on the full 1064 vocabulary with 
            Watch-Lookup).
    }
    \label{tab:mouthingthres}
\end{table}

\noindent\textbf{Effect of the sliding window stride.}
As explained in Sec.~\ref{subsection:implementation},
at test time, we extract features from the continuous signing
sequence using a sliding window approach with 1 frame as the stride parameter.
In Tab.~\ref{tab:stride}, we investigate the effect of the stride parameter.
Our window size is 16 frames, i.e., the number of input frames for the I3D
feature extractor.
A standard approach when extracting features from longer videos
    is to use a sliding window with 50\% overlap (i.e., stride of 8 frames).
    However, this means the temporal resolution of the search space is
    reduced by a factor of 8, and a stride of 8 may skip the most discriminative
moment since a sign duration is typically between 
7-13 frames (but can be shorter)~\cite{pfister2013large} in continuous signing video.
    In Tab.~\ref{tab:stride}, we see that we can gain a significant localisation
    improvement by computing the similarities more densely, e.g., stride of 4 frames
    may be sufficiently dense. In our experiments, we use stride 1.

We refer to
\if\sepappendix1{Appendix~B}
\else{Appendix~\ref{app:sec:additionalexp}}
\fi
for additional ablations.

\begin{table}[t]
    \setlength{\tabcolsep}{8pt}
    \centering
    \resizebox{0.6\linewidth}{!}{
        \begin{tabular}{lcc}
            \toprule
            Stride & mAP & R@5 \\
            \midrule
            16 & 31.96 & 38.98 \\
            8 & 38.46 & 47.38 \\
            4 & 44.92 & 54.65 \\
            2 & \textbf{45.39} & \textbf{55.63} \\
            1 & 43.65 & 53.03 \\
            \bottomrule
        \end{tabular}
    }
    \caption{\textbf{Stride parameter of sliding window:} A small stride
    at test time, when extracting embeddings from the continuous signing video,
    allows us to temporally localise the signs more precisely. The window size is
    16 frames and the typical co-articulated sign duration is 7-13 frames at 25 fps.
    (testing 1064-class model trained with Watch-Lookup)
    }
    \label{tab:stride}
\end{table}

\begin{figure*}[t]
  \centering
  \includegraphics[width=0.97\textwidth]{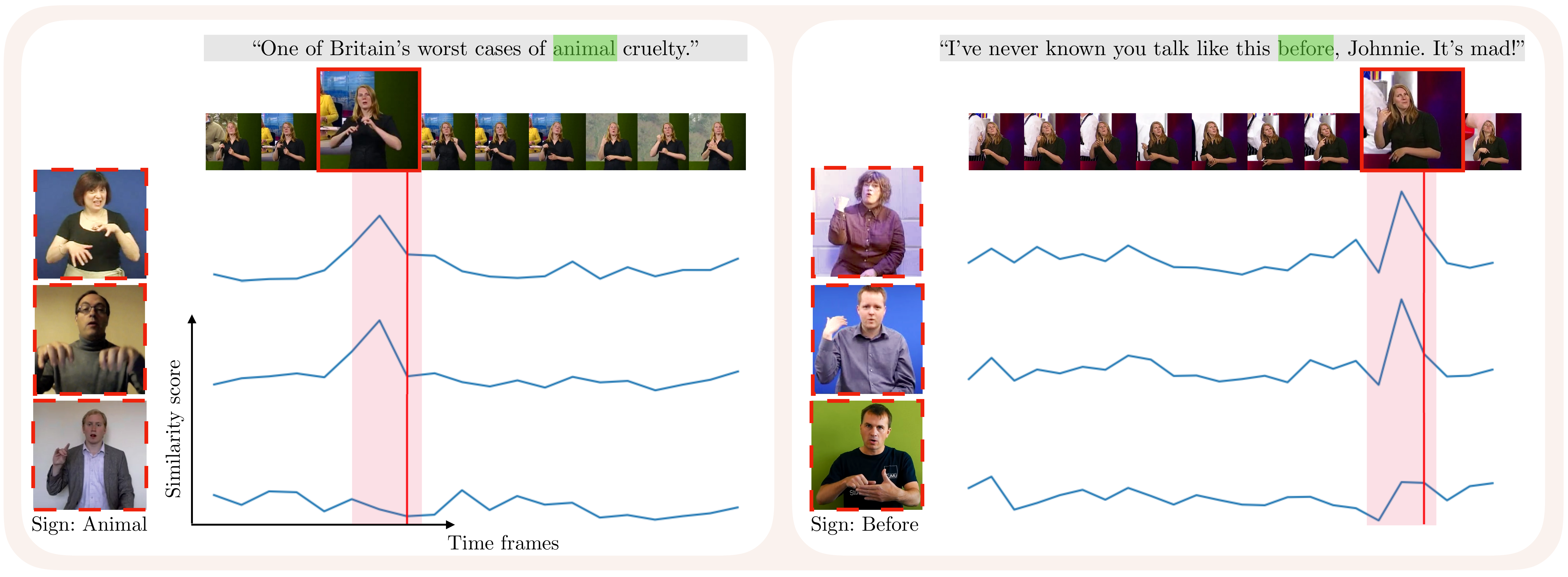}
  \caption{\textbf{Sign variant identification:} We plot the similarity scores between 
      \bslonek{} test clips and \bsldict{} variants of the sign ``animal'' (left) and 
      ``before'' (right) over time.
      A high similarity occurs for the first two rows, where the \bsldict{} examples match 
      the variant used in \bslonek{}.
      The labelled mouthing times from 
      \cite{Albanie20} are shown by red vertical lines and approximate windows for signing times are shaded. 
      Note that neither the mouthing annotations (ground truth)
          nor the dictionary spottings provide the \textit{duration} of the sign,
          but only a point in time where the response is highest.
          The mouthing peak (red vertical line) tends to appear at the end of the sign (due to the use of LSTM
          in visual keyword spotter).
          The dictionary peak (blue curve) tends to appear in the middle of the sign.
}
  \label{fig:variants}
\end{figure*}
\begin{figure*}
  \centering
  \includegraphics[width=0.98\textwidth]{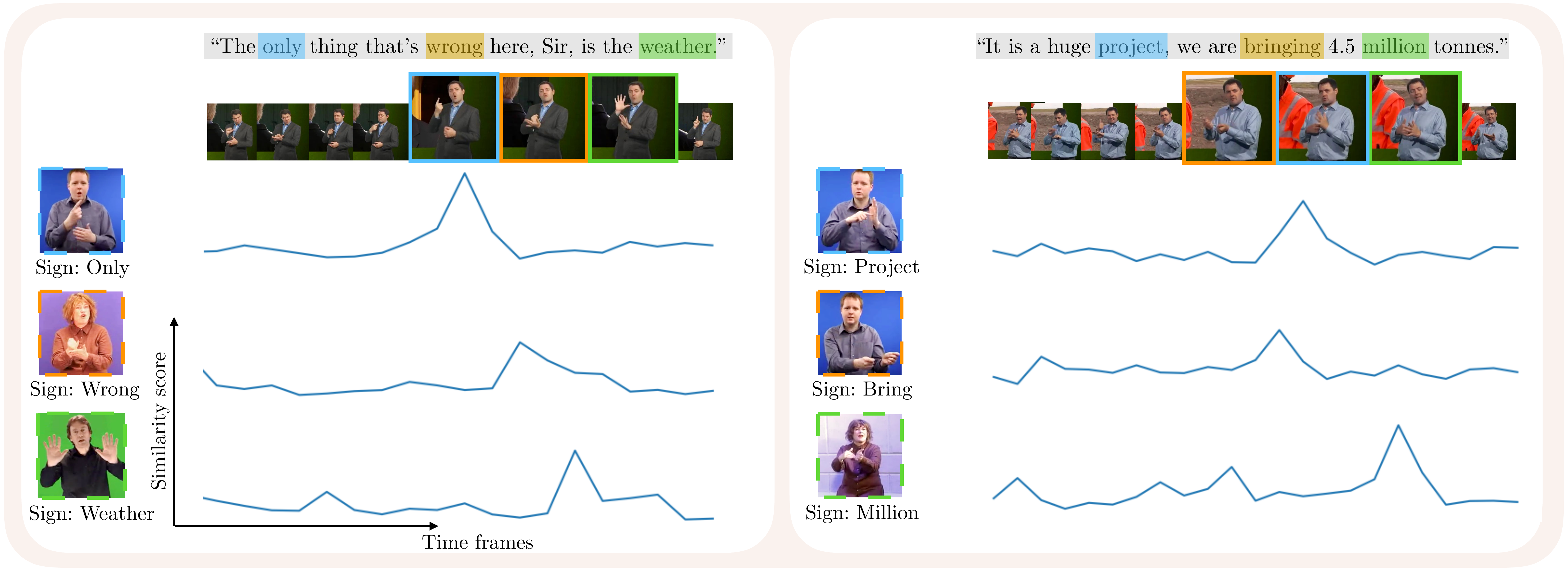}
  \caption{\textbf{Densification:} We plot the similarity scores between \bslonek{} test 
      clips and \bsldict{} examples over time, by querying only the words in the subtitle. 
      We visually inspect the quality of the dictionary spottings with
          which we obtain cases of multiple words per subtitle spotted.
      The predicted locations of the signs correspond to the peak similarity scores.
      Note that unlike in Fig.~\ref{fig:variants}, 
          we cannot overlay the ground truth since the annotations using the mouthing cues
          are not dense enough to provide ground truth sign locations for 3 words per subtitle.
  }
  \label{fig:densification}
\end{figure*}

\subsection{Applications}\label{subsection:applications}
In this section, we investigate three applications of our sign spotting method.

\noindent\textbf{Sign variant identification.} We show the ability of our model to spot
specifically which variant of the sign was used. In Fig.~\ref{fig:variants}, we observe 
high similarity scores when the variant of the sign matches in both \bslonek{} and 
\bsldict{} videos.  Identifying such sign variations allows a better understanding of 
regional differences and can potentially help standardisation efforts of BSL.

\noindent\textbf{Dense annotations.} We demonstrate the potential of our model to obtain dense annotations
on continuous sign language video data. Sign spotting through the use of sign dictionaries 
is not limited to mouthings as in~\cite{Albanie20} and therefore is of
great importance to scale up datasets for learning more robust sign language models.
In Fig.~\ref{fig:densification}, we show qualitative examples
of localising multiple signs in a given sentence in \bslonek{}, where we only query
the words that occur in the subtitles, reducing the search space. In fact, if we assume the 
word to be known, we obtain {83.08\%} sign localisation accuracy on \bslonek{} with our 
best model. This is defined as the number of times the maximum similarity occurs within 
-20/+5 frames of the end label time provided by~\cite{Albanie20}. %

\begin{figure*}[t]
  \centering
  \includegraphics[width=0.98\textwidth]{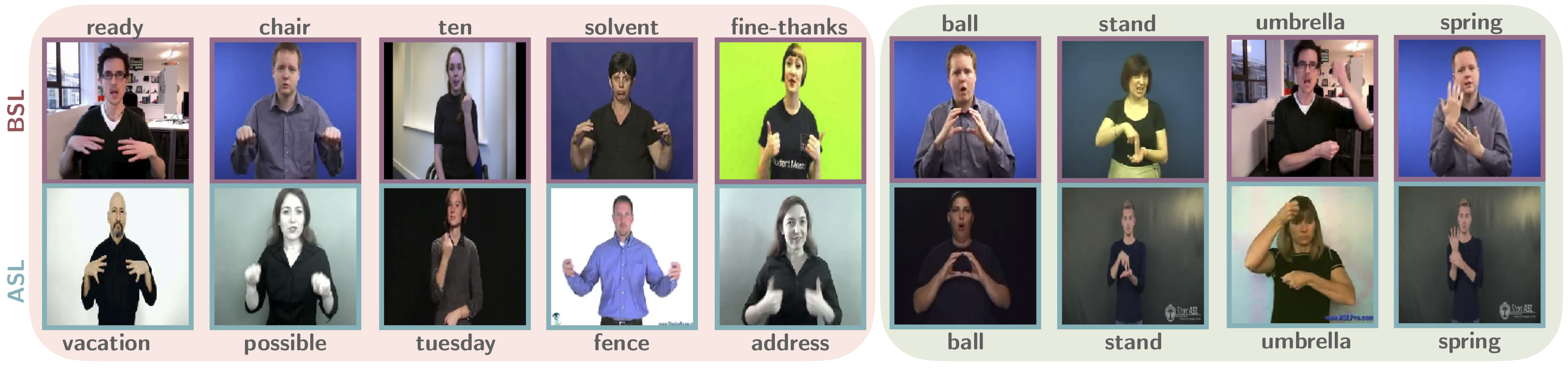}
  \caption{\textbf{``Faux amis'' in BSL/ASL:}  Same/similar manual features for different 
      English words (left), as well as for the same English words (right), are identified 
      between \bsldict{} and WLASL isolated sign language datasets.}
  \label{fig:fauxamis}
\end{figure*}

\noindent\textbf{``Faux Amis''.} There are works investigating lexical similarities between 
sign languages manually~\cite{SignumMcKee2000,Aldersson2007}. We show qualitatively the 
potential of our model to discover similarities, as well as ``faux-amis" between different 
sign languages, in particular between British (BSL) and American (ASL) Sign Languages. We 
retrieve nearest neighbors according to visual embedding similarities between \bsldict{} 
which has a 9K vocabulary and WLASL~\cite{Li19wlasl}, an ASL isolated sign language 
dataset with a 2K vocabulary. We provide some examples in Fig.~\ref{fig:fauxamis}.
We automatically identify several signs with similar manual features
some of which correspond to different meanings in English (left),
as well as same meanings,
such as ``ball'', ``stand'', ``umbrella'' (right).

\begin{figure*}
    \centering
    \begin{tabular}{ccc}
        & \qquad Yield for 1K query vocabulary & \qquad Yield for 9K query vocabulary \\
        \includegraphics[width=0.32\textwidth]{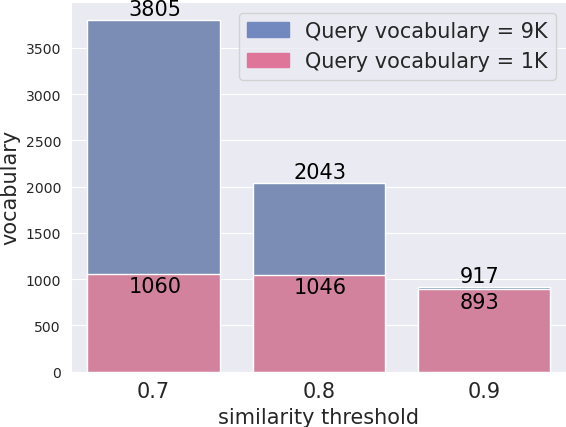} &
        \includegraphics[width=0.32\textwidth]{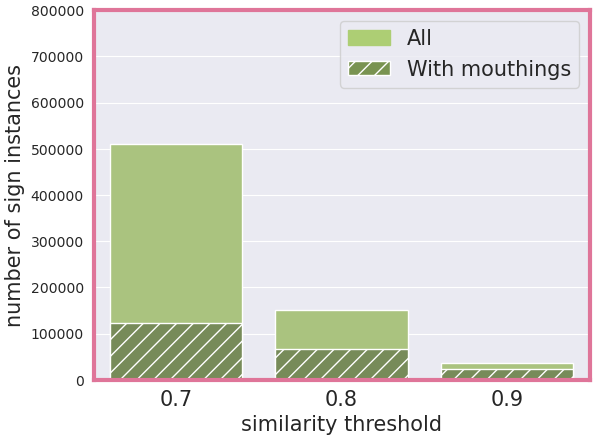} &
        \includegraphics[width=0.32\textwidth]{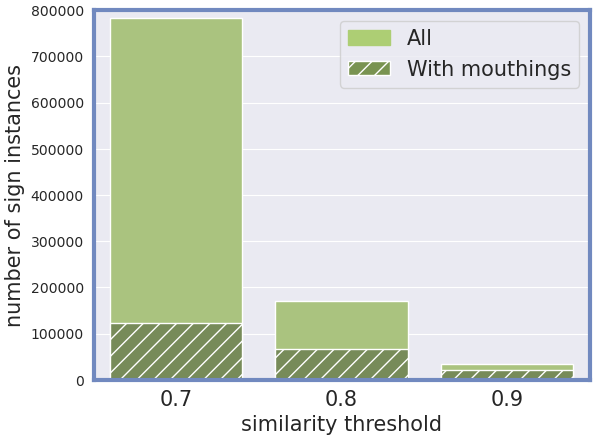} \\
    \end{tabular}
    \caption{\textbf{Statistics on the yield from the automatic annotations:}
            We plot the vocabulary size (left) and the number of localised sign instances
            (middle) and (right) over several similarity thresholds for the new automatic
            annotations in the training set that we obtain through dictionaries.
            While we obtain a large number of localised signs (783K at 0.7 threshold)
            for the full 9K vocabulary, in our recognition experiments we use a subset of 510K
            annotations that correspond to the 1K vocabulary. To approximately quantify
            the amount of annotations that represent duplicates from those found through mouthing cues,
            we count those localisations for which the same keyword exists for mouthing annotations
            within the same search window. We observe that the majority of the
            annotations are new (783K vs 122K).
    }
    \label{fig:stats}
\end{figure*}

\subsection{Sign language recognition}\label{subsection:recognition}

As demonstrated qualitatively in Sec.~\ref{subsection:applications}, we can reliably obtain
automatic annotations using our sign spotting technique when the search space is reduced to candidate
words in the subtitle. A natural way to exploit our method
is to apply it on the BSL-1K \textit{training} set in conjunction with the weakly-aligned
subtitles to collect new localised sign instances. This allows us to train a sign recognition model:
in this case,
to \textit{retrain} the I3D architecture from~\cite{Albanie20} which was previously supervised only
with signs localised through mouthings.

\noindent \textbf{BSL-1K automatic annotation.} Similar to our previous work 
    using mouthing cues \cite{Albanie20},
where words in the subtitle were queried within a neighborhood
around the subtitle timestamps, we query each subtitle word
if they fall within a predefined set of vocabulary.
In particular, we query words and phrases from the 9K \bsldict{} vocabulary if they occur 
in the subtitles. To determine whether a query from the dictionary occurs
in the subtitle, we apply several checks. We look for the original word or phrase
as it appears in the dictionary, as well as its text-normalised form (e.g., ``20'' becomes
``twenty''). For the subtitle, we look for its original, text-normalised, and lemmatised forms.
Once we find a match between any form of the dictionary text and any form of the subtitle text,
we query the dictionary video feature within the search window
in the continuous video features.
We use search windows of $\pm$4 seconds padding around the subtitle
timestamps. We compute the similarity between the continuous signing search window 
and each of the dictionary
variants for a given word: 
we record the frame location of maximum similarity for all variants and choose the best 
match as the one with highest similarity score.
The final sign localisations are obtained
by filtering the peak similarity scores to those above 0.7 threshold -- resulting in a 
vocabulary of 4K signs -- 
and taking 32 frames centered around the peak location.
Fig.~\ref{fig:stats}
summarises several statistics computed over the training set.
We note that sign spotting with dictionaries (D) is more effective
than with mouthing (M) in terms of the yield (510K versus 169K localised signs).
Since, D can include duplicates from M, we further report the number
of instances for which a mouthing spotting for the same keyword query
exists within the same search window. We find that the majority of our
D spottings represent new, not previously localised instances (see Fig.~\ref{fig:stats} right).

\noindent \textbf{BSL-1K sign recognition benchmark.}
We use the BSL-1K manually verified recognition test set with 2K samples~\cite{Albanie20},
which we denote with \textbf{\testrec{}},
and significantly extend it to 37K samples as \textbf{\testrecnew{}}.
We do this by (a) running our dictionary-based sign spotting technique
on the BSL-1K test set and (b) verifying the predicted sign instances
with human annotators using the VIA 
tool~\cite{Dutta19a} as in~\cite{Albanie20}.
Our goal in keeping these two divisions is three-fold:
    (i) \testrec{} is the result of annotating ``mouthing'' spottings
    above 0.9 confidence, which means the models can largely rely on mouthing
    cues to recognise the signs. The new \testrecnew{} annotations have both ``mouthing''
    (10K) and ``dictionary'' (27K) spottings. The dictionary annotations are the result
    of annotating dictionary spottings above 0.7 confidence from this work;
    therefore, models are required to recognise the signs even in the absence
    of mouthing, reducing the bias towards signs with
    easily spotted mouthing cues.
    (ii) \testrecnew{} spans a much larger fraction of the training
    vocabulary as seen in Tab.~\ref{tab:datasets}, with~950 out of 1,064 sign classes
    (vs only 334 classes in the original benchmark \testrec{} of~\cite{Albanie20}).
    (iii) We wish
    to maintain direct comparison to our previous work~\cite{Albanie20}; therefore,
    we report on both sets in this work.

\begin{table}
    \setlength{\tabcolsep}{1pt}
    \centering
    \resizebox{0.999\linewidth}{!}{
        \begin{tabular}{lr|cccc|cccc}
            \toprule
            & & \multicolumn{4}{c}{\testrec{}~\cite{Albanie20}} & \multicolumn{4}{|c}{\testrecnew{}}\\
            \midrule
            & & \multicolumn{4}{c}{2K inst. / 334 cls.} & \multicolumn{4}{|c}{37K inst. / 950 cls.}\\
            \midrule
            & & \multicolumn{2}{c}{per-instance} & \multicolumn{2}{c}{per-class} & \multicolumn{2}{|c}{per-instance} & \multicolumn{2}{c}{per-class} \\
            Training & \#ann. & top-1 & top-5 & top-1 & top-5 & top-1 & top-5 & top-1 & top-5 \\
            \midrule
            M~\cite{Albanie20}$\mathsection$ & 169K  & 76.6 & 89.2 & 54.6 & 71.8 & 26.4 & 41.3	& 19.4 & 33.2 \\
            D & 510K & 70.8 & 84.9 & 52.7 & 68.1 & 60.9 & 80.3 & 34.7 & 53.5 \\
            M+D & 678K & \textbf{80.8} & \textbf{92.1} & \textbf{60.5} & \textbf{79.9} & \textbf{62.3} & \textbf{81.3} & \textbf{40.2} & \textbf{60.1} \\
            \bottomrule
        \end{tabular}
    }
    \vspace{2pt}
    \caption{\textbf{An improved I3D sign recognition model:}
            We find
            signs via automatic dictionary spotting (D), significantly
            expanding the training and testing data obtained from mouthing cues by~\cite{Albanie20} (M).
            We also significantly expand the test set by manually
            verifying these new automatic annotations from the test partition
            (\testrec{} vs \testrecnew{}).
            By training on the extended M+D data, we obtain state-of-the-art results,
            outperforming the previous work of~\cite{Albanie20}.
            $\mathsection$The slight improvement in the performance of \cite{Albanie20}
            over the original results reported in that work is due to 
            our denser test-time averaging when
            applying sliding windows (8-frame vs 1-frame stride).
    }
    \label{tab:recognition}
\end{table}

\noindent \textbf{Comparison to prior work.}
In Tab.~\ref{tab:recognition}, we compare three I3D models trained on mouthing annotations (M),
dictionary annotations (D) , and their combination (M+D).
First, we observe that D-only model significantly outperforms M-only model
on \testrecnew{} (60.9\% vs 26.4\%),
while resulting in lower performance on \testrec{} (70.8\% vs 76.6\%).
This may be due to the strong bias towards mouthing cues in the small test set \testrec{}.
Second, the benefits of combining annotations from both can be seen
in the sign classifier trained using 678K automatic annotations. This obtains 
state-of-the-art performance on \testrec{}, as well as the more challenging test set \testrecnew{}.
All three models in the table (M, D, M+D) are pretrained on Kinetics~\cite{Carreira2017}, followed
by video pose distillation as described in~\cite{Albanie20}. We observed
no improvements when initialising M+D training from M-only pretraining.

Our results can be interpreted as bootstrapping from an initial model,
which has access to a large audio-visual training set with mouthing annotations.
The M recognition model has distilled this information while incorporating manual patterns.
The Watch-Read-Lookup framework has mainly relied on these mouthing locations
to learn matching with dictionary samples. The D recognition model 
is the result of this series of annotation expansion. 
The final recognition model therefore exploits multiple sources
of supervision. We refer to our recent work \cite{varol21bslattend}
for a complementary way of expanding the automatic annotations.
There, we introduce an attention-based sign localisation where
the localisation ability emerges from a sequence prediction task.

\begin{table}
    \setlength{\tabcolsep}{4pt}
    \centering
    \resizebox{0.99\linewidth}{!}{
        \begin{tabular}{lr|cccc}
            \toprule
            & & \multicolumn{4}{|c}{\testrecnew{}}\\
            \midrule
            & & \multicolumn{2}{c}{per-instance} & \multicolumn{2}{c}{per-class} \\
            Training & \#ann. & top-1 & top-5 & top-1 & top-5 \\
            \midrule
            D$_{.9}$ (sim $\geq0.9$) & 36K & 28.4 & 44.8 & 15.8 & 28.8 \\
            D$_{.8}$ (sim $\geq0.8$) & 152K & 60.6 & 78.0 & \textbf{35.6} & 52.6 \\
            D$_{.7}$ (sim $\geq0.7$) & 510K & \textbf{60.9} & \textbf{80.3} & 34.7 & \textbf{53.5} \\
            \midrule
            D$_{.7}$ (no padding) & 236K & 56.5 & 75.9 & 28.7 & 45.4 \\
            \bottomrule
        \end{tabular}
    }
    \vspace{2pt}
    \caption{\textbf{Recognition ablations:} We train on a portion of the
            automatic annotations obtained through dictionaries. We filter the localisations
            which have more than a similarity threshold: 0.7, 0.8, 0.9. We find that lower
            threshold results in a larger training and increased performance. On the other hand,
            restricting the annotations to only those which fall within the subtitle timestamps
            without temporal padding for the search window reduces the accuracy.
    }
    \label{tab:recognitionablation}
\end{table}

\begin{table}
    \setlength{\tabcolsep}{4pt}
    \centering
    \resizebox{0.99\linewidth}{!}{
        \begin{tabular}{lr|cccc}
            \toprule
            & & \multicolumn{4}{|c}{BOBSL}\\
            \midrule
            & & \multicolumn{2}{c}{per-instance} & \multicolumn{2}{c}{per-class} \\
            Training & \#ann. & top-1 & top-5 & top-1 & top-5 \\
            \midrule
            M$_{.8}$ & 154K & 39.0 & 55.1 & 43.0 & 66.3 \\ %
            M$_{.5}$ & 502K & 39.6 & 55.2 & 44.3 & 66.3 \\ %
            \midrule
            D$_{.9}$ & 9K & 24.0 & 39.5 & 4.5 & 9.6 \\ %
            D$_{.8}$ & 272K & 63.2 & 79.0 & 25.7 & 41.2 \\ %
            D$_{.75}$ & 727K & 64.8 & 80.8 & 29.1 & 44.7 \\ %
            \midrule
            M$_{.8}$+D$_{.8}$ & 426K & \textbf{75.8} & 92.4 & 50.5 & \textbf{77.6} \\ % 426429
            M$_{.5}$+D$_{.75}$ & 1.2M & \textbf{75.8} & \textbf{92.5} & \textbf{51.1} & 77.1 \\ % 1228788
            \bottomrule
        \end{tabular}
    }
    \vspace{2pt}
    \caption{\textbf{Recognition results on the BOBSL dataset:}
            We report 2281-way classification performance
            on the 25,045 manually verified test signs of BOBSL.
            We train on different training sets obtained with
            automatic annotations through mouthing (M) and dictionaries (D)
            at different thresholds.
            Combining
            two sets of annotations yields the best result,
            even without lowering the thresholds.
    }
    \label{tab:bobsl}
\end{table}

\noindent\textbf{Sign recognition ablations.}
In Tab.~\ref{tab:recognitionablation}
we provide further ablations for training the
recognition models based on automatic dictionary spotting
annotations. In particular, we investigate
(i) the similarity threshold that determines 
the amount of training data, as well as the noise, and
(ii) no padding versus $\pm$4-second padding to the subtitle locations
when defining the search window.
We see in Tab.~\ref{tab:recognitionablation} that
filtering the sign annotations with a high threshold
such as 0.9, denoted with D$_{.9}$,
drastically reduces the training size (from
510K to 36K) which in return results in poor recognition performance.
The accuracy with D$_{.7}$ is slightly above that of D$_{.8}$.
Moreover, both the performance and the training size
decreases if we restrict the sign annotations
to those which fall within the subtitle timestamps,
i.e., no temporal padding in the search window when
applying sign spotting. We retain a similarity
threshold of 0.7 and a $\pm$4-second padding for our final model.

\subsection{Results on the BOBSL dataset}
\label{subsection:bobsl}

\noindent\textbf{BOBSL~\cite{Albanie2021bobsl}}
    is a dataset similar to \bslonek{}; however, unlike \bslonek{}, BOBSL is publicly available.
    The dataset consists of 1,400 hours of BSL-interpreted
    BBC broadcast footage accompanied by written English subtitles.
    We repeated our sign spotting techniques on this data using
    mouthing and dictionary cues in combination with subtitles.
    Keyword spotting with mouthing
    follows our previous work~\cite{Albanie20} and obtains
    502K sign localisations over 0.5 confidence (M$_{0.5}$).
    Sign spotting with
    dictionaries is similar to the procedure described in
    Sec.~\ref{subsection:recognition}, %
    resulting in 727K sign localisations over 0.75
    similarity (D$_{0.75}$).

    In Tab.~\ref{tab:bobsl}, we present sign recognition
    results using these automatic annotations for classification
    training over a vocabulary of 2281 categories.
    The BOBSL test set contains 25,045 manually verified
    signs obtained through both types of spotting techniques.
    We experiment with various sets of annotations for training.
    We observe that mouthing (M) and dictionary (D) spottings
    are complementary.
    Similar to Tab.~\ref{tab:recognitionablation},
    we find that lowering the similarity threshold improves
    the performance for D-only training.
    However, when combined into a significantly larger
    training set (i.e., a total of 1.2 million clips with
    low thresholds), this improvement disappears (75.8\% top-1 accuracy for both).

\begin{figure}
    \includegraphics[width=0.45\textwidth]{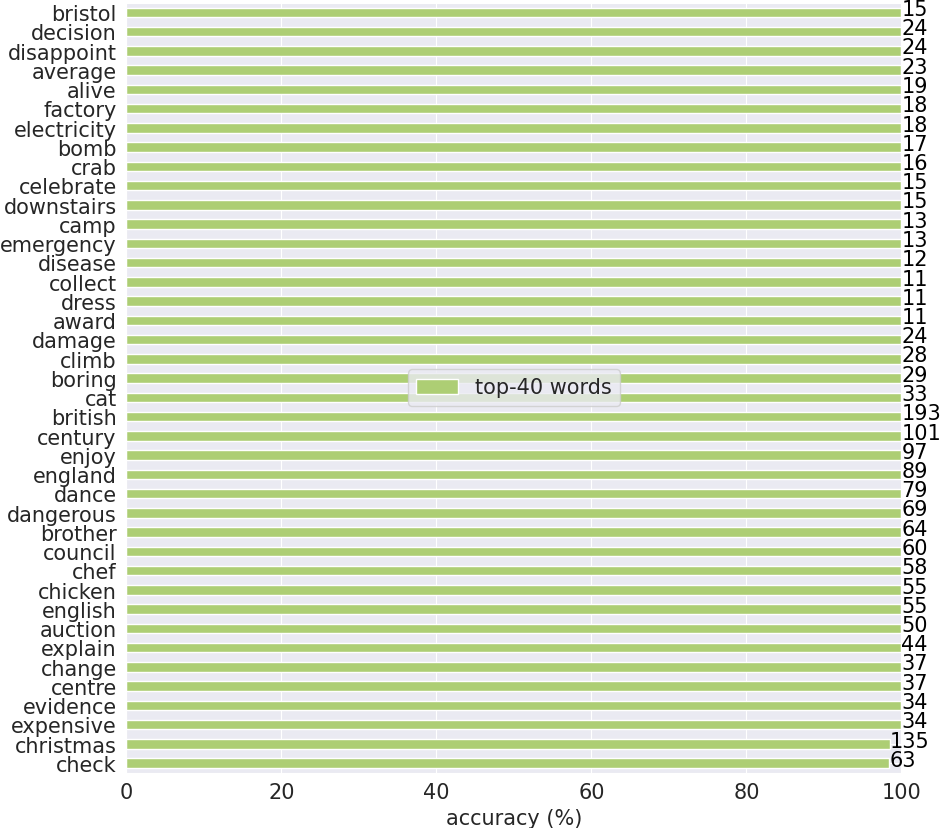}
    \includegraphics[width=0.45\textwidth]{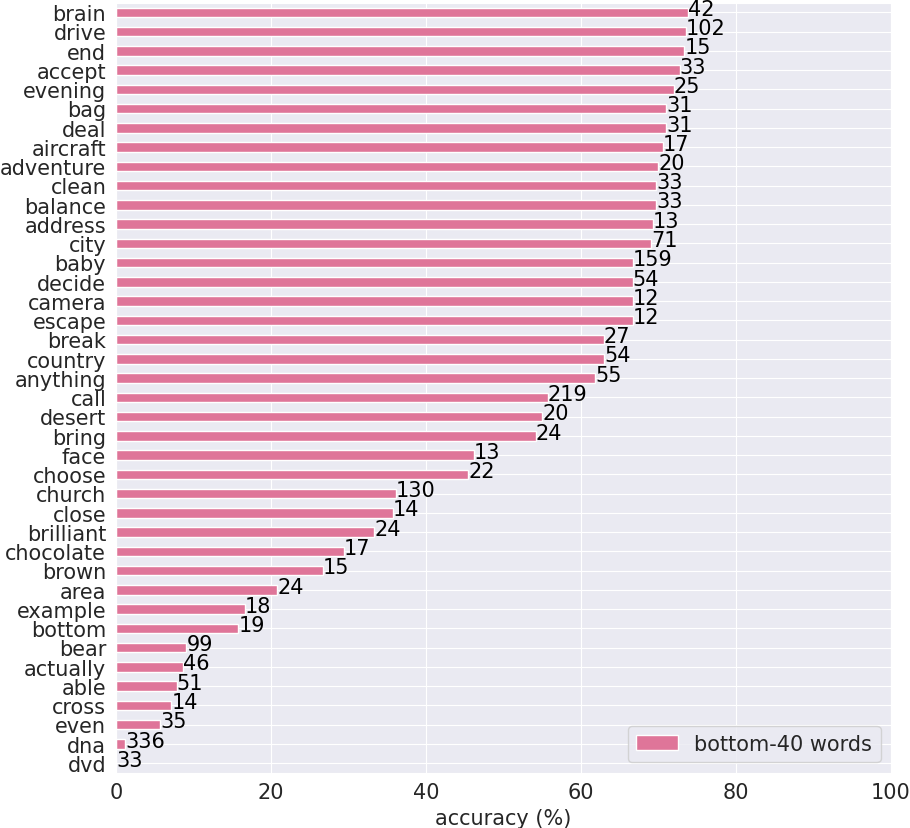}
    \caption{\textbf{Error analysis:}
            We plot the per-word verification accuracies
            on BOBSL for top- and bottom-40 words sorted
            according to accuracies. Each bar also shows
            the total number of manual verifications performed
            for the corresponding word. While many words have
            close to 100\% accuracy, certain words fail drastically
            such as `dvd' and `dna', mainly due to being fingerspelled
            signs. Other words such as `even' and `able' may have
            different meanings depending on context. Note that
            the query word here is obtained from subtitles.
    }
    \label{fig:erroranalysis1}
\end{figure}

\begin{figure}
    \includegraphics[width=0.49\textwidth]{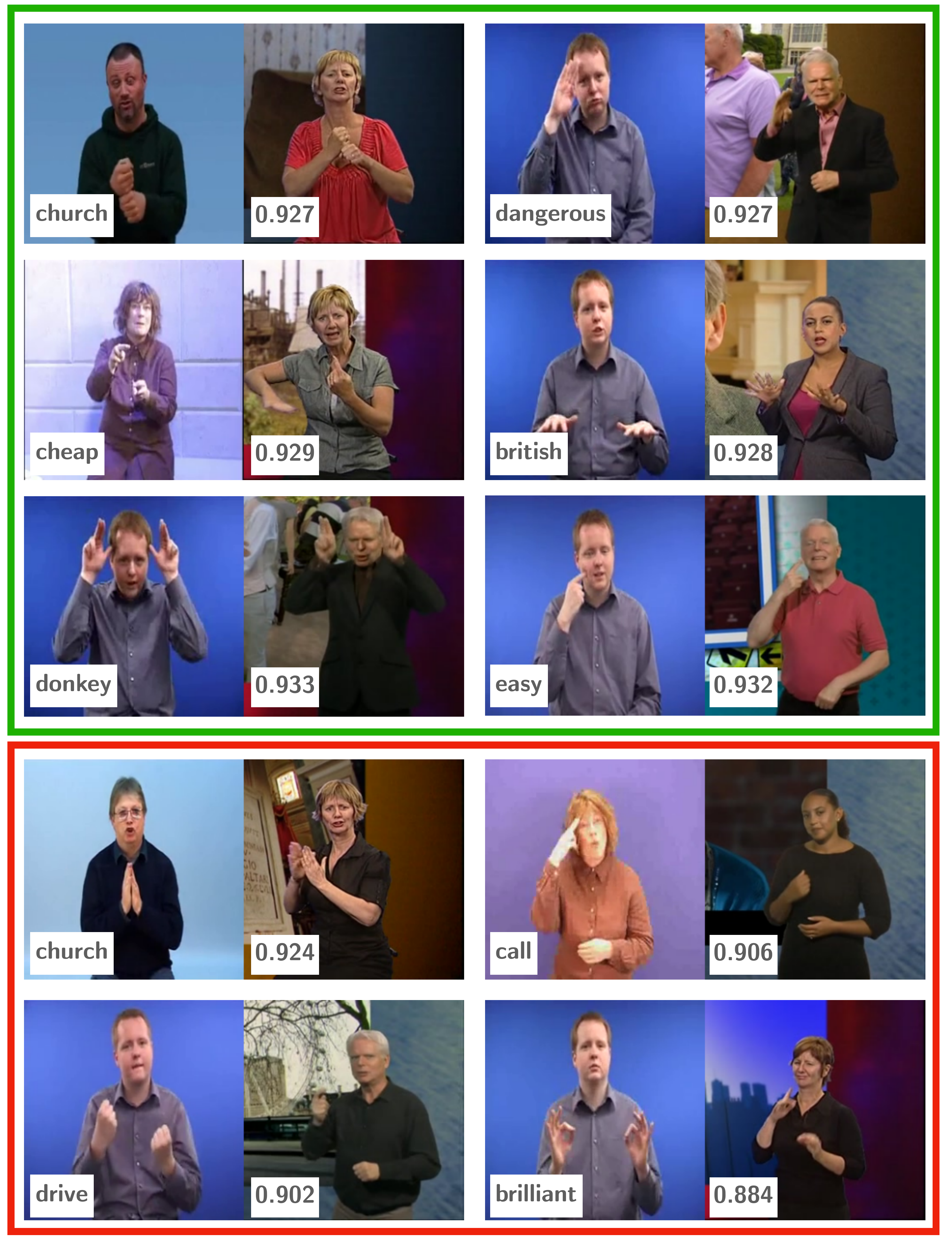}
    \caption{\textbf{Qualitative analysis:} We
            visualise sample spotting results from the manually
            verified set of BOBSL that obtain the highest similarity scores
            (overlaid on the spotted frame).
            Note that the query word is obtained
            from the subtitles. Top and bottom blocks represent success
            and failure cases, respectively. We notice weak hand shape and motion similarities
            in the failing examples. The figure only shows the middle frame for
            each video; therefore, we provide video visualisations at
            \url{https://www.robots.ox.ac.uk/~vgg/research/bsldict/viz/viz_bobsl_dicts.html}.
    }
    \label{fig:erroranalysis2}
\end{figure}

\subsection{Limitations.}
\label{subsection:limitations}

In this section, we investigate failure modes of our sign
spotting mechanism, in particular by using the data
obtained through manual verifications.
More specifically, we make statistics
from 10K annotations on the BOBSL test set
that were obtained via dictionary spotting through
subtitles. From these, 76\% were marked as correct.
In Fig.~\ref{fig:erroranalysis1},
we present a breakdown for per-word
accuracy to check whether certain signs fail
more than others. We note two main failure modes:
(i) fingerspelled words (e.g., `dvd', `dna')
are difficult for the model, perhaps due to
sparse frame sampling from long dictionary videos, (ii)
common words such as `even' and `able' 
may have context-dependent meanings; querying
these words due to %
occurrence in subtitles lead to false positives.

In Fig.~\ref{fig:erroranalysis2},
we further visualise samples from this
manually verified set of spottings.
We focus on cases where high similarities
occur and group the examples into
success (top) and failure (bottom) cases. Within failures,
we observe weak hand shape and motion similarities.
As previously mentioned, this might be due
to querying a word for which
a sign correspondence does not exist
within the temporal search window.
Future work may develop a mechanism to determine
which words to query from the subtitle to ensure
correspondence with a sign, so that the problem
only becomes localisation.

\section{Conclusions} \label{sec:conclusions}
We have presented an approach to spot signs
in continuous sign language videos using visual sign dictionary videos, and have shown the 
benefits of leveraging multiple supervisory signals
available in a realistic setting: (i) sparse annotations in continuous signing (in our 
case, from mouthings),
(ii) accompanied with subtitles, and (iii) a few dictionary samples per word
from a large vocabulary. We employ multiple-instance contrastive learning to incorporate
these signals into a unified framework. 
We finally propose several potential applications of sign spotting and 
demonstrate its ability to scale up sign language datasets for training strong 
sign language recognition models.

\bigskip
\noindent\textbf{Acknowledgements.}
This work was supported by EPSRC grant ExTol and a Royal Society Research Professorship.
The authors would to like thank Abhishek Dutta, A.~Sophia Koepke, Andrew Brown, Necati Cihan Camg\"oz,
Neil Fox, Joon Son Chung,
Bencie Woll,
and Hannah Bull
 for their help.
The authors are also grateful to Daniel Mitchell who made \url{signbsl.com} webpage available.
SA would like to thank Z. Novak and S. Carlson for enabling his contribution.

\bibliographystyle{splncs04}
\bibliography{shortstrings,vgg_local,references}

\clearpage
\noindent \textbf{\large APPENDIX}
\bigskip
\renewcommand{\thefigure}{A.\arabic{figure}}
\setcounter{figure}{0} 
\renewcommand{\thetable}{A.\arabic{table}}
\setcounter{table}{0}

\renewcommand{\thesection}{\Alph{section}}
\setcounter{section}{0}

This appendix provides additional qualitative (Sec.~\ref{app:sec:qualitative}) and 
experimental results (Sec.~\ref{app:sec:additionalexp}),
as well as detailed explanations of the training of our Watch-Read-Lookup framework 
(Sec.~\ref{app:sec:details}).

\section{Qualitative Results}\label{app:sec:qualitative}
Please watch our %
video in the project webpage\footnote{\url{https://www.robots.ox.ac.uk/~vgg/research/bsldict/}}
to see qualitative results of our model in action. We  illustrate the sign spotting task, 
as well as the specific applications considered in the main paper: sign variant 
identification, densification of annotations, and ``faux amis'' identification between languages.

\section{Additional Experiments}\label{app:sec:additionalexp}
In this section, we present complementary experimental results
to the main paper.
We report
the effect of class-balancing (Sec.~\ref{app:subsec:balancing}),
domain-specific layers (Sec.~\ref{app:subsec:domainspecific}),
language-aware negative sampling (Sec.~\ref{app:subsec:languageaware}),
and the trunk network architecture
(Sec.~\ref{app:subsec:s3d}).

\begin{table}[h!]
    \setlength{\tabcolsep}{8pt}
    \centering
    \resizebox{0.6\linewidth}{!}{
        \begin{tabular}{llcc}
            \toprule
            Class-balancing & Batch size & mAP & R@5 \\
            \midrule
            \xmark & 512 & 41.65 & 54.73 \\
            \xmark & 1024 & 42.07 & 54.25 \\
            \xmark & 2048 & 43.14 & 54.28 \\
            \midrule
            \cmark & 512 & 43.65 & 53.03 \\
            \cmark & 1024 & 43.55 & 54.20 \\
            \bottomrule
        \end{tabular}
    }
    \caption{\textbf{Class-balancing:}
    In the main paper, we class-balance our minibatches by including
    one sample per word from the labelled continuous sequences, thus maximizing the number of negatives
    within a batch.
    Here, we investigate removing such class-balancing constraint.
    In that case, we make sure we do not mark samples
    with the same labels as negatives, instead we discard them.
    We experiment with various batch sizes, also going beyond
    the total number of classes (2048). We observe that
    the performance is not significantly affected
    by these changes.
    (training on the full 1064 vocabulary with Watch-Lookup)
    }
    \label{app:tab:balancing}
\end{table}

\subsection{Class-balanced sampling}\label{app:subsec:balancing}
As described in the main paper, we construct each batch by maximizing the number
of negative pairs. To this end, we include one labelled sample per word when sampling
continuous sequences, i.e., class-balancing the minibatches. Thus,
all but one of the labelled samples in the batch can be used as negatives
for a given dictionary bag corresponding to a labelled sample.
Note that this approach limits the batch size to be less than or equal to the
number of sign classes.
Tab.~\ref{app:tab:balancing} experiments with the sampling strategy.
We observe that the performance is not significantly
different with/without class-balanced sampling
for various batch sizes.

\subsection{Domain-specific layers}\label{app:subsec:domainspecific}
As noted in the main paper, the videos from the continuous signing and from the dictionaries
differ significantly, e.g., continuous signing data is faster than the dictionary signing, and is co-articulated whereas the dictionary has isolated signs.
Given such a domain gap, we explore whether it is beneficial to learn
domain-specific MLP layers: one for the continuous, and one for the dictionary.
Tab.~\ref{app:tab:domainspecific} presents a comparison between
domain-specific layers versus shared parameters. We do not observe any
gains from such separation. Therefore, we keep a single MLP for both domains
for simplicity.

\begin{table}[t]
    \setlength{\tabcolsep}{8pt}
    \centering
    \resizebox{0.6\linewidth}{!}{
        \begin{tabular}{lcc}
            \toprule
            Domain-specific layers  & mAP & R@5 \\
            \midrule
            \cmark & 43.58 & 53.54 \\
            \xmark & 43.65 & 53.03 \\
            \bottomrule
        \end{tabular}
    }
    \caption{\textbf{Domain-specific layers:} We experiment with separating the MLP layers
    to be specific to the continuous and isolated domains. We do not observe
    any significant difference in performance and therefore adopt a shared MLP
    for simplicity in all experiments. (Training on the full 1064 vocabulary with Watch-Lookup)
    }
    \label{app:tab:domainspecific}
\end{table}

\subsection{Language-aware negative sampling}\label{app:subsec:languageaware}
Working with a large vocabulary of words brings the additional
challenge of handling synonyms. We consider two types of similarities.
First, two different categories in the \bsldict{} sign dictionary may belong
to the same sign category if the corresponding English words are synonyms.
Second, the meta-data we have collected with the \bsldict{} dataset
provides similarity labels between sign categories, which may be used
to group certain signs.
In this work, we have largely ignored this issue by associating
each sign to a single word. This results in constructing
negative pairs for two identical signs such as `happy' and `content'.
Here, we explore whether it is beneficial to discard
such pairs during training, instead of marking them as negatives.
Tab.~\ref{app:tab:languageaware}
reports the results. We observe marginal gains with discarding synonyms.
However, given the insignificant difference, we
do not make such separation in other experiments for simplicity.

\begin{table}[t]
    \setlength{\tabcolsep}{8pt}
    \centering
    \resizebox{0.7\linewidth}{!}{
        \begin{tabular}{lcc}
            \toprule
            Negative sampling  & mAP & R@5 \\
            \midrule
            Discarding English synonyms & 43.27 & 54.24 \\
            Discarding Sign synonyms & 45.03 & 54.19 \\
            Keeping all & 43.65 & 53.03 \\
            \bottomrule
        \end{tabular}
    }
    \caption{\textbf{Language-aware negative sampling:} We explore the use of external knowledge
    such as English synonyms or the meta-data of the dictionary denoting similar sign categories.
    We experiment with discarding such similar word pairs, excluding them from both
    positive and negative pairs. The last row instead marks any pair
    as negative if their corresponding words are not identical. We observe only marginal
    gains with the use of external knowledge about the languages. (Training on the full 1064 vocabulary with Watch-Lookup)
    }
    \label{app:tab:languageaware}
\end{table}

\subsection{Trunk network architecture: S3D vs I3D} \label{app:subsec:s3d}
As shown in Tab~\ref{app:tab:s3d},
we compare two popular architectures
for computing video representations. We have used
I3D~\cite{Carreira2017} in all our experiments.
Here, we also train a 1064-way classification
with the S3D architecture~\cite{XieS3D}
on \bslonek{} as in \cite{Albanie20} for sign
language recognition. We do not observe
improvements with S3D (in practice we found that it overfit the training set to a greater degree); therefore, we use
an I3D trunk. Note that the hyperparameters
(e.g., learning rate)
are tuned
for I3D and kept the same for S3D.

\begin{table}
    \setlength{\tabcolsep}{8pt}
    \centering
    \resizebox{0.7\linewidth}{!}{
        \begin{tabular}{l|cccc}
            \toprule
            & \multicolumn{2}{c}{per-instance} & \multicolumn{2}{c}{per-class} \\
            Training data & top-1 & top-5 & top-1 & top-5  \\
            \midrule
            S3D & 64.76	& 81.88	& 46.27 & 63.71 \\
            I3D~\cite{Albanie20} & \textbf{75.51} & \textbf{88.83} & \textbf{52.76} & \textbf{72.14} \\
            \bottomrule
        \end{tabular}
    }
    \caption{
    \textbf{Trunk network architecture:} We compare
    I3D~\cite{Carreira2017} with the S3D~\cite{XieS3D} architecture
    for the task of sign language recognition,
    in a comparable setup to~\cite{Albanie20}.
    We use the last 20 frames before the mouthing
    annotations with confidence above 0.5. We do not
    obtain gains with the S3D architecture; therefore,
    we use I3D in all the experiments to compute video
    features.
    }
    \label{app:tab:s3d}
\end{table}

\section{Training Details}\label{app:sec:details}
In this section, we cover architectural details (Sec.~\ref{app:subsec:arch}),
a detailed formulation of our positive/negative
bag sampling strategy (Sec.~\ref{app:subsec:math}) and a brief description of the infrastructure used to perform the experiments in the main paper (Sec.~\ref{app:subsec:infra}).

\begin{figure}[t]
    \centering
    \includegraphics[width=.49\textwidth]{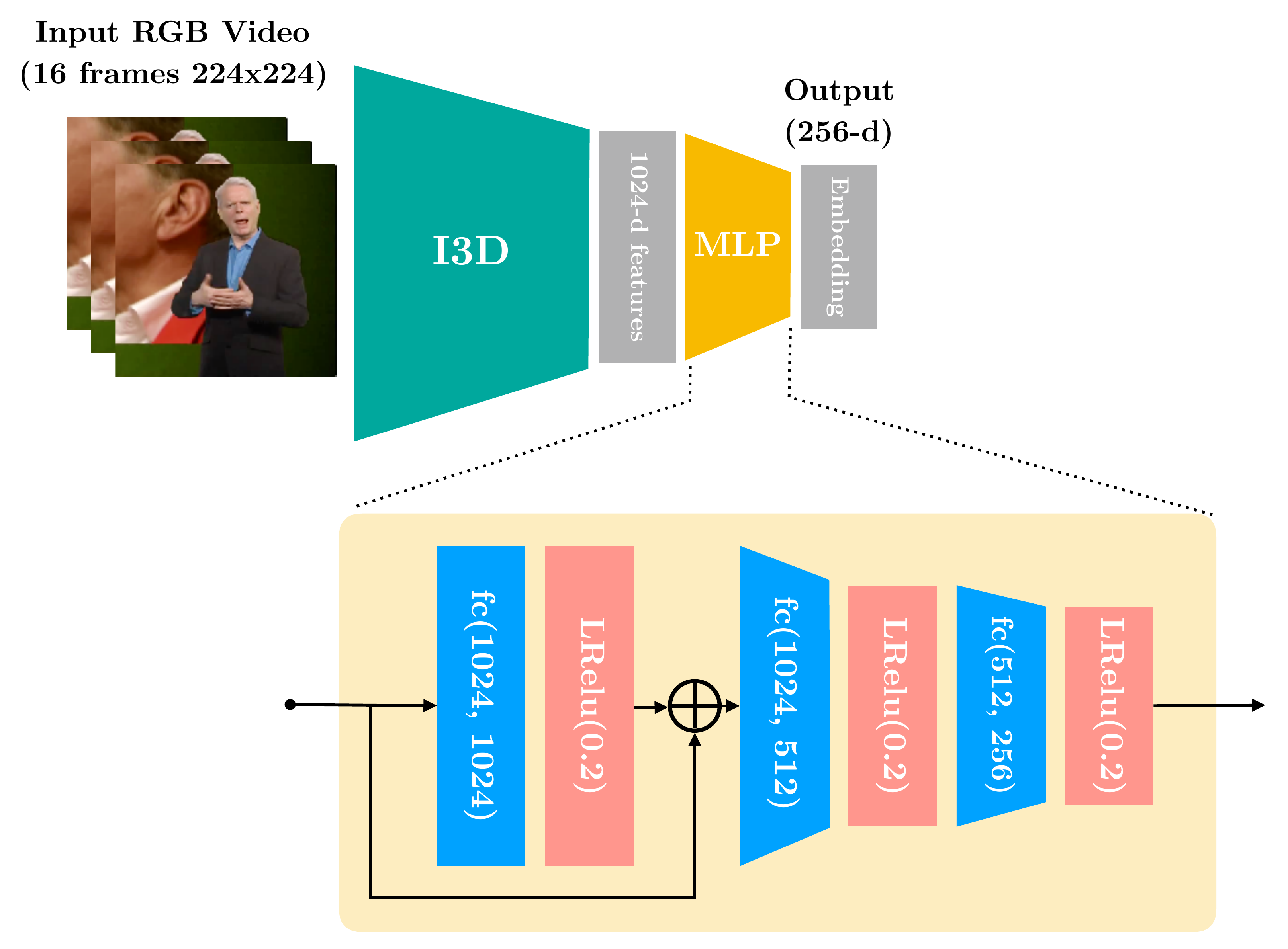}
    \caption{\textbf{MLP architecture:} We detail the layers of our embedding architecture.
    We freeze the I3D trunk and use it as a feature extractor.
    We only train the MLP layers with our loss formulation in the proposed framework. The same layers (and parameters) are used both for the dictionary video inputs and the continuous signing video inputs.}
    \label{app:fig:arch}
\end{figure}

\subsection{Architectural details}\label{app:subsec:arch}
As explained in the main paper, our sign embeddings correspond
to the output of a two-stage architecture: (i) an I3D trunk, and (ii) a three-layer MLP.
We first train the I3D on both labelled continuous video clips
and the dictionary videos jointly. We then freeze the I3D trunk
and use it as a feature extractor. We only train the MLP layers with our loss formulation in the Watch-Read-Lookup framework. \\

\noindent\textbf{I3D trunk.} %
We first train the I3D parameters only with the \bslonek{} annotated clips
that have mouthing confidences more than 0.5. For 1064-class training,
we use the publicly available model from \cite{Albanie20}; for 800-class training, we perform
our own training, also first pretraining with pose distillation.

We then \textit{re-initialise the batch normalization layers}
(as noted in
\if\sepappendix1{Sec.~2 of the main paper).}
\else{Sec.~\ref{sec:related} of the main paper).}
\fi
We fine-tune the model jointly on \bslonek{} annotated clips (the ones
with mouthing confidence more than 0.8)
and \bsldict{} samples. The sampling frequency for the two data sources are balanced.
In the I3D classification pretraining phase, we treat each dictionary video
independently with its corresponding label.
We observe that the 1064-way classification performance
on the \textit{training} dictionary videos remain at 48.09\% per-instance top-1
accuracy without the batch normalization re-initialization, as opposed
to 78.94\%. We also experimented with domain-specific batch normalization layers~\cite{chang2019domain},
but the training accuracy for the dictionary videos was still low (62.73\%).

As detailed in
\if\sepappendix1{Sec.~3.2 of the main paper,}
\else{Sec.~\ref{subsection:implementation} of the main paper,}
\fi
we subsample the dictionary
videos to roughly match their speed to the continuous signing videos.
This subsampling includes \textit{a random shift and a random fps}. We observe
a decrease of 6.68\% in the training dictionary classification accuracy
if we instead sample 16 consecutive frames from the original
temporal resolution, which is not sufficient to capture the full extent of
a sign because one dictionary video is 56 frames on average. \\

\noindent\textbf{MLP.} Fig.~\ref{app:fig:arch} illustrates
the layers considered for our MLP architecture. It consists of 3 fully connected
layers with LeakyRelu activations between them. The first linear layer also has a
residual connection on the 1024-dimensional input features. We then reduce
the dimensionality gradually to 512 and 256 for efficient training and testing. \\

\subsection{Positive/Negative bag sampling formulations}\label{app:subsec:math}

In the main paper, we described two approaches for sampling positive/negative MIL bags in
\if\sepappendix1{Sec.~3.1.}
\else{Sec.~\ref{subsection:mil}.}
\fi
Due to space constraints, the sampling mechanisms were described at a high-level.  Here, we provide more precise definitions of each bag.  In addition to the set notation below, we include in the code release, the loss implementation as a PyTorch~\cite{NEURIPS2019_9015} function in \texttt{loss/loss.py}, together with a sample
input ({\texttt{loss/sample\_inputs.pkl}}) comprising embedding outputs from the MLP
for continuous and dictionary videos. %

As noted in the main paper, we do not have access to positive pairs because: (1) for the segments of videos in $\mathcal{S}$ that are annotated (i.e. $(x_k, v_k) \in \mathcal{M}$), we have a set of potential sign variations represented in the dictionary (annotated with the common label $v_k$), rather than a single unique sign; (2) since $\mathcal{S}$ provides only weak supervision, even when a word is mentioned in the subtitles we do not know where it appears in the continuous signing sequence (if it appears at all). 
These ambiguities motivate a Multiple Instance Learning~\cite{dietterich1997solving} (MIL) objective. Rather than forming positive and negative pairs, we instead form positive \textit{bags} of pairs, $\mathcal{P}^{\text{bags}}$, in which we expect at least one segment from a video from $\mathcal{S}$ (or a video from $\mathcal{M}$ when labels are available) and a video $\mathcal{D}$ to contain the same sign, and negative bags of pairs, $\mathcal{N}^{\text{bags}}$, in which we expect no pair of video segments from $\mathcal{S}$ (or $\mathcal{M}$) and $\mathcal{D}$ to contain the same sign.  To incorporate the available sources of supervision into this formulation, we consider two categories of positive and negative bag formations, described next. Each bag is formulated as a set of paired indices---the first value indexes into the collections of continuous signing videos (either $\mathcal{S}$ or $\mathcal{M}$, depending on context) and the second value indexes into the set of dictionary videos contained in $\mathcal{D}$. \\

\noindent \textbf{Watch and Lookup: using sparse annotations and dictionaries}. In the first formulation, \textit{Watch-Lookup}, we only make use of $\mathcal{D}$ and $\mathcal{M}$ (and not $\mathcal{S}$) to learn the data representation $f$.  We define positive bags in two ways: (1) by anchoring on the labelled segment  

\begin{align}
    \mathcal{P}_{\text{watch,lookup}}^{\text{bags(seg)}} = &\{\{i\} \times B_i: (x_i^\mathcal{M}, v_i^\mathcal{M}) \in \mathcal{M}, \nonumber \\
    &(x_j^\mathcal{D}, v_j^\mathcal{D}) \in \mathcal{D}, B_i = \{j : v_j^\mathcal{D} =  v_i^\mathcal{M}\}\}
\end{align}

\begin{figure*}[t]
    \centering
    \includegraphics[width=\textwidth]{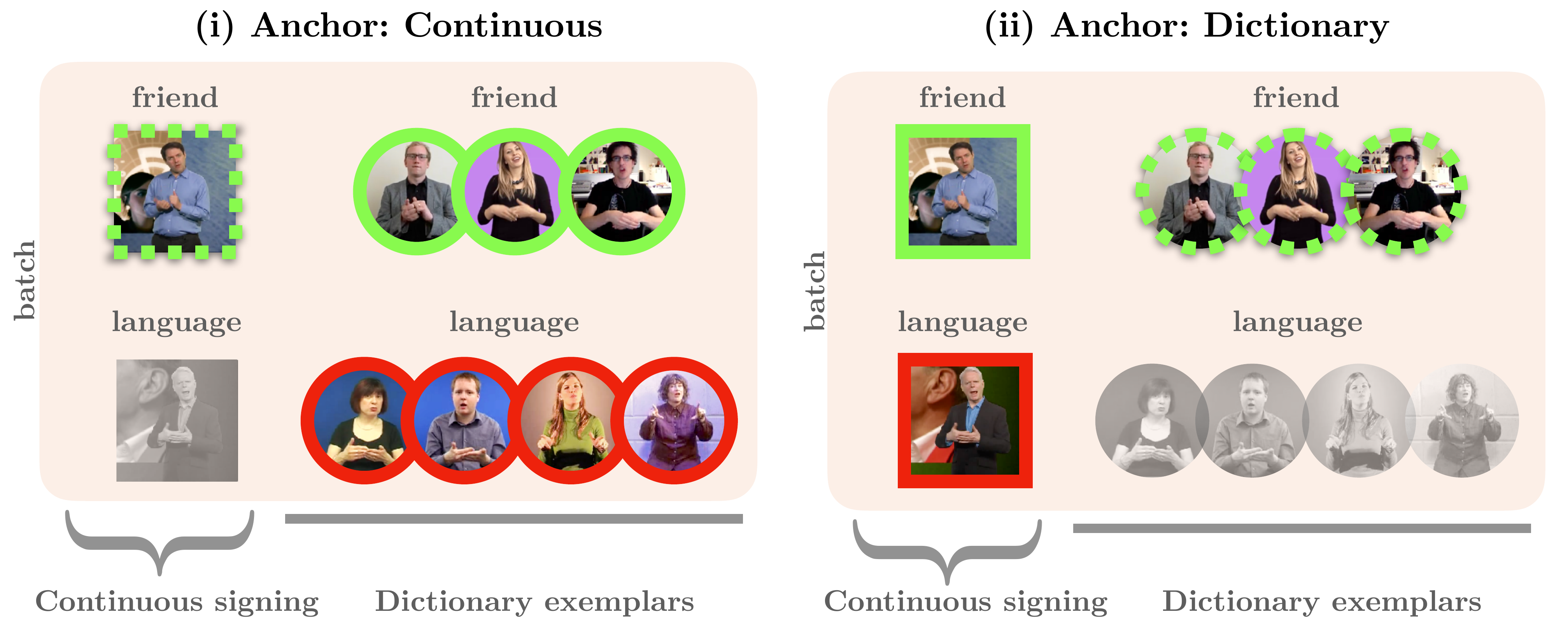}
    \caption{\textbf{Watch-Lookup:} We illustrate the batch formation and positive/negative sampling
    for the simplified version of our framework which
    is not using the subtitles, but only performing
    Watch-Lookup.
    We define two sets of positive/negative pairs, anchoring at a different position in each case.
    Anchor is denoted with dashed lines, positive samples
    with solid green, negative samples with solid red lines. Gray samples are discarded.
    (i) anchors at a labelled continuous video, making the dictionary samples for the labelled word a positive bag, and all other dictionary samples in the batch a negative bag.
    (ii) anchors at a bag of dictionary samples, making the corresponding continuous labelled video positive,
    and all others in the batch negatives.
    We refer to Fig~\ref{app:fig:sampling_mil2nce} for the illustration
    of our Watch-Read-Lookup extension.
    }
    \label{app:fig:sampling_milnce}
\end{figure*}

\noindent i.e. each bag consists of a labelled temporal segment and the set of sign variations of the corresponding word in the dictionary (illustrated in Fig.~\ref{app:fig:sampling_milnce} (i), top row), or by (2) anchoring on the dictionary samples that correspond to the labelled segment, to define a second set $\mathcal{P}_{\text{watch,lookup}}^{\text{bags(dict)}}$, which takes a mathematically identical form to $\mathcal{P}_{\text{watch,lookup}}^{\text{bags(seg)}}$ (i.e. each bag consists of the set of sign variations of the word in the dictionary that corresponds to a given labelled temporal segment, illustrated in Fig.~\ref{app:fig:sampling_milnce} (ii), top row). The key assumption in both cases is that each labelled segment matches \textit{at least one} sign variation in the dictionary. Negative bags can be constructed by (1) anchoring on labelled segments and selecting dictionary examples corresponding to different words (Fig.~\ref{app:fig:sampling_milnce} (i), red examples); (2) anchoring on the dictionary set for a given word and selecting labelled segments of a different word  (Fig.~\ref{app:fig:sampling_milnce} (ii), red example).
These sets manifest as

\begin{align}
    \mathcal{N}_{\text{watch,lookup}}^{\text{bags(seg)}} = & \{\{i\} \times B_i: (x_i^\mathcal{M}, v_i^\mathcal{M}) \in \mathcal{M}, \nonumber \\
    & (x_j^\mathcal{D}, v_j^\mathcal{D}) \in \mathcal{D}, B_i = \{j : v_j^\mathcal{D} \neq v_i^\mathcal{M} \}\}
\end{align}

\noindent for the former and as

\begin{align}
    &\mathcal{N}_{\text{watch,lookup}}^{\text{bags(dict)}} = \{A_i \times B_i: \nonumber \\ &A_i = \{l: x_l, x_i \subseteq x_k, (x_k, s_k) \in \mathcal{S}, x_l \cap x_i = \emptyset\}, \nonumber \\
    & B_i = \{j : v_j^\mathcal{D} \neq v_i^\mathcal{M} \}, (x_i^\mathcal{M}, v_i^\mathcal{M}) \in \mathcal{M}, (x_j^\mathcal{D}, v_j^\mathcal{D}) \in \mathcal{D}
    \}\}.
\end{align}

\noindent for the latter. The complete set of positive and negative bags is formed via the unions of these collections:

\begin{align}
\mathcal{P}_{\text{watch,lookup}}^{\text{bags}} \triangleq \mathcal{P}_{\text{watch,lookup}}^{\text{bags(seg)}} \cup \mathcal{P}_{\text{watch,lookup}}^{\text{bags(dict)}}
\end{align}

\noindent and

\begin{align}
    \mathcal{N}_{\text{watch,lookup}}^{\text{bags}} \triangleq \mathcal{N}_{\text{watch,lookup}}^{\text{bags(seg)}} \cup \mathcal{N}_{\text{watch,lookup}}^{\text{bags(dict)}}.
\end{align}

\begin{figure*}
    \centering
\subfloat[
\textbf{Input:} We illustrate an example minibatch formation
for our Watch-Read-Lookup framework. We sample continuous
videos with only one labelled segment, which we refer to as the `foreground' word (e.g., \textit{friend}, \textit{language}). Each continuous video has a subtitle, which we use to sample additional words for which we do not have continuous signing labels, (`background' words), e.g. \textit{name} and \textit{what} for \textit{``what is your friend's name?''}. We sample all the dictionary videos corresponding to these words. Each word has multiple dictionary instances grouped into overlapping circles.
]
    {\makebox[12cm][c]{
    \includegraphics[width=0.9\textwidth]{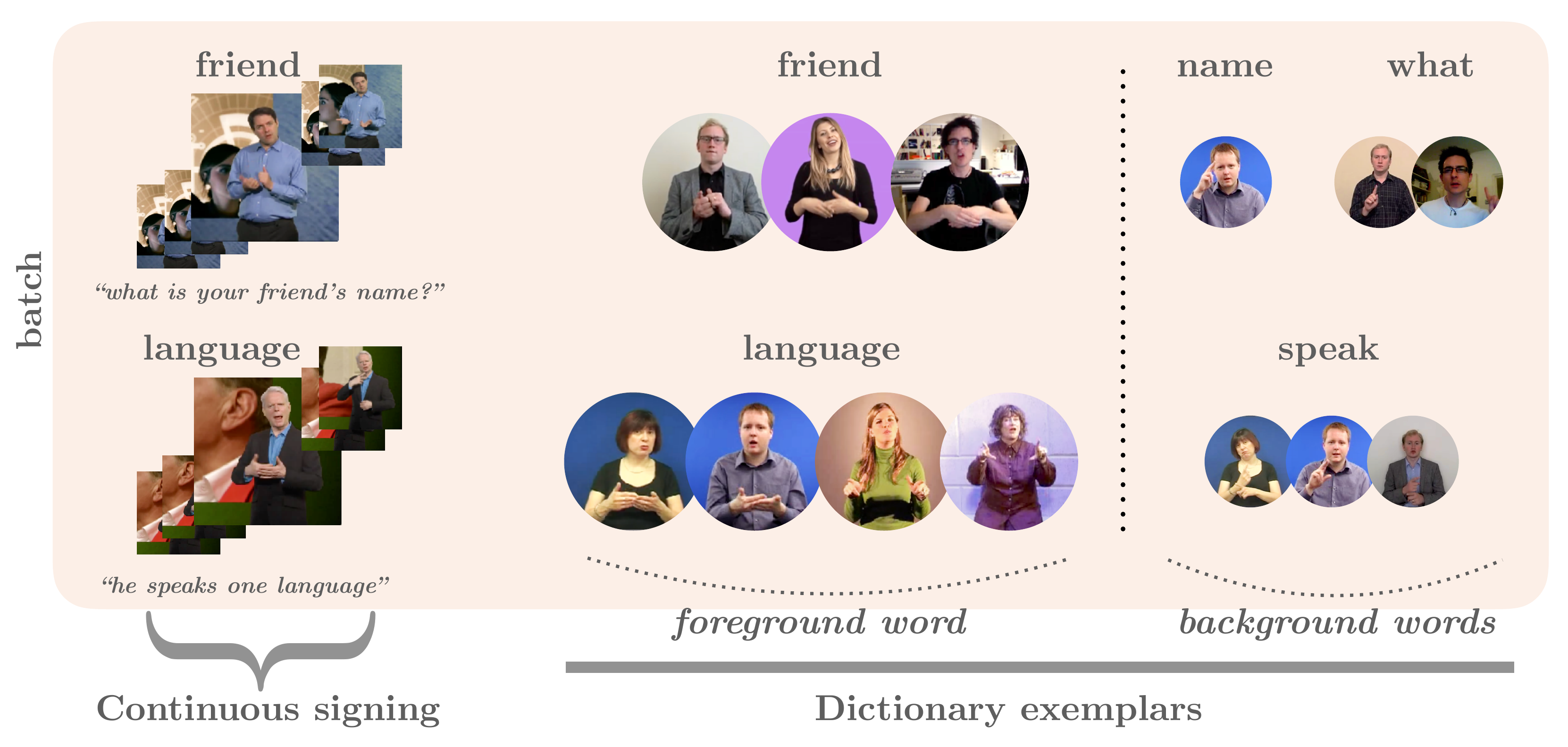}}} \\
\subfloat[
\textbf{Sampling positive/negative pairs:}
We anchor at 4 different positions within the batch to determine the pairs. Anchors are denoted with dashed lines, positive samples with solid green, negative samples with solid red lines. Gray samples are discarded.
For example, (iii) anchoring at the continuous background
marks the dictionary video for \textit{name} positive, because
it appears in the subtitle, but it is not within the annotated
temporal window. All other dictionary samples \textit{friend, language, speak} become negative to this anchor.
We repeat this for each dictionary background, i.e.,
marking \textit{what} as positive.
See text for detailed explanations on each case.
We also provide a %
video animation at our project page
to show all possible positive/negative pairs
for cases (i) to (iv).
]{
    \includegraphics[width=\textwidth]{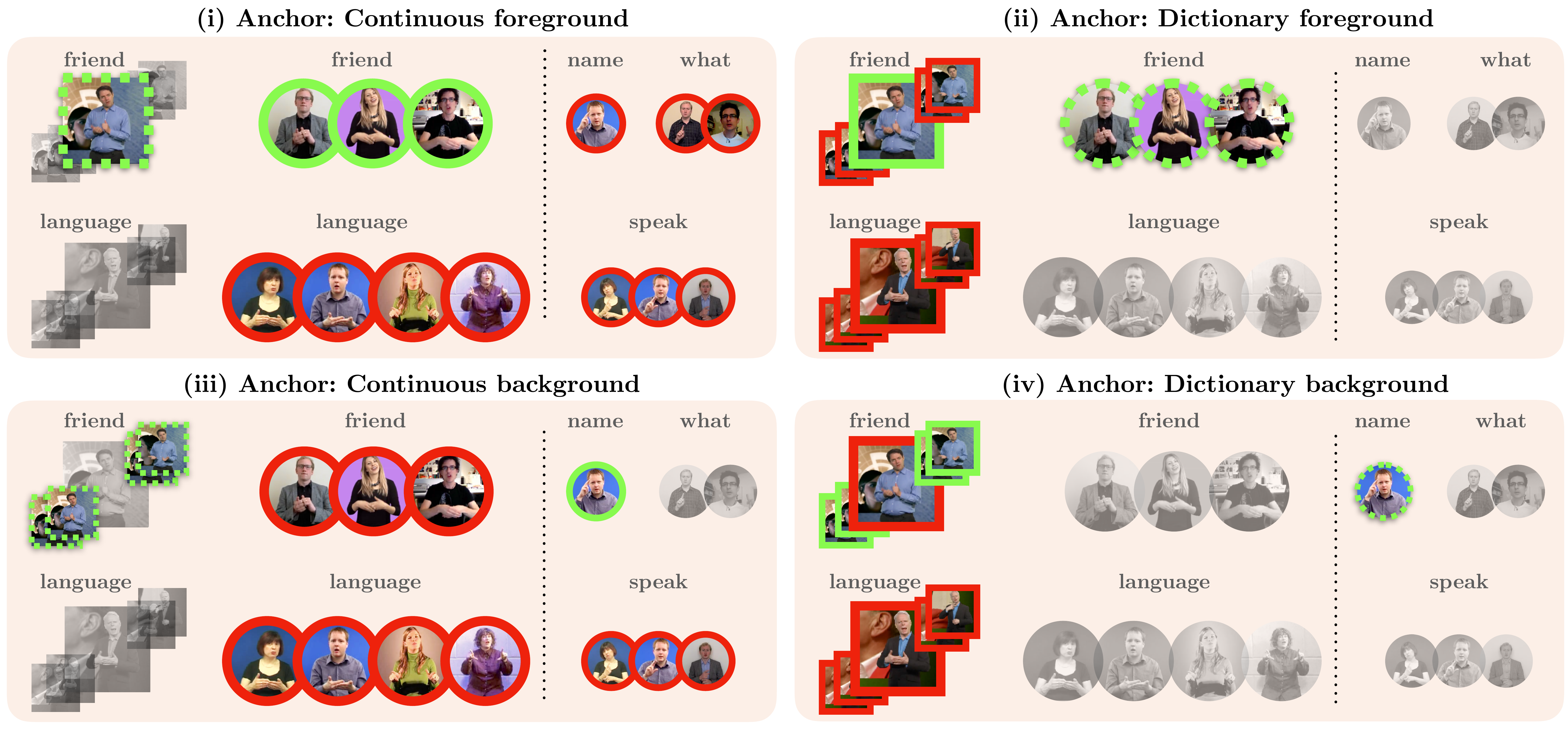}}
    \caption{\textbf{Watch-Read-Lookup in detail.}
    }
    \label{app:fig:sampling_mil2nce}
\end{figure*}

\noindent \textbf{Watch, Read and Lookup}. The \textit{Watch-Lookup} bag formulation defined above has a significant limitation: the data representation, $f$, is not encouraged to represent signs beyond the initial vocabulary represented in $\mathcal{M}$.  We therefore look at the subtitles present in $\mathcal{S}$ (which contain words beyond $\mathcal{M}$) in addition to $\mathcal{M}$ to construct bags. To do so, we introduce an additional piece of terminology---when considering a subtitled video for which only one segment is labelled, we use the term \say{foreground} to refer to the subtitle word that corresponds to the label, and \say{background} for words which do not possess labelled segments in the video. Similarly to \textit{Watch-Lookup}, we can construct positive bags,  $\mathcal{P}_{\text{watch,lookup}}^{\text{bags}}$ (Fig.~\ref{app:fig:sampling_mil2nce} (i) and (ii), top rows) which correspond to the use of foreground subtitle words.  However, these can now by extended by (a) anchoring on a background segment in the continuous footage and find candidate matches in the dictionary among all possible matches for the subtitles words (Fig.~\ref{app:fig:sampling_mil2nce} (iii), top row) and (b) anchoring on dictionary entries for background subtitle words (Fig.~\ref{app:fig:sampling_mil2nce} (iv), top row). Formally, let Tokenize$(\cdot): \mathcal{S} \rightarrow \mathcal{V}_\mathfrak{L}$ denote the function which extracts words from the subtitle that are present in the vocabulary: Tokenize$(s) \triangleq \{ w \in s: w \in \mathcal{V}_\mathfrak{L}\}$. Then define background segment-anchored positive bags as: 
\nopagebreak

\begin{align}
&\mathcal{P}_{\text{watch,read,lookup}}^{\text{bags(seg-back)}} = \{\{i\} \times B_i: \nonumber \\
& \quad \quad \quad \exists (x_k, s_k) \in \mathcal{S} \text{ s.t } x_i \subseteq x_k, (x_j^\mathcal{D}, v_j^\mathcal{D}) \in \mathcal{D}, \nonumber \\
& \quad \quad \quad B_i = \{j : v_j^\mathcal{D} \in \text{Tokenize}(s_k)\}, (x_i, v_i) \notin \mathcal{M}\}\}
\end{align}

\noindent i.e. each bag contains a background segment from the continuous signing which is paired with all dictionary segments whose labels match any token from the corresponding subtitle sentence (visualised as the top row of Fig.~\ref{app:fig:sampling_mil2nce} (iii)). Next, we define dictionary-anchored positive background bags as follows:
\begin{align}
    & \mathcal{P}_{\text{watch,read,lookup}}^{\text{bags(dict-back)}} = \{A_i \times B_i: (x_i^\mathcal{D}, v_i^\mathcal{D}) \in \mathcal{D}, \nonumber \\
    & \quad \quad \quad A_i = \{j : v_i^\mathcal{D} \in \text{Tokenize}(s_k), (x_k, s_k) \in \mathcal{S}, x_j \subseteq x_k, \nonumber \\
    & \quad \quad \quad (x_j, v_j) \notin \mathcal{M}\},  B_i = \{l: v_l^\mathcal{D} = v_i^\mathcal{D}\}\}
\end{align}

\noindent i.e. the bags contain all pairwise combinations of dictionary entries for a given word and segments in continuous signing whose subtitle contains that background word (visualised as top row of Fig.~\ref{app:fig:sampling_mil2nce} (iv)). 
We combine these bags with the \textit{Watch-Lookup} positive bags to maximally exploit the available supervisory signal for positives:

\begin{align}
\mathcal{P}_{\text{watch,read,lookup}}^{\text{bags}} = & \, \, \mathcal{P}_{\text{watch,lookup}}^{\text{bags}} \cup \mathcal{P}_{\text{watch,read,lookup}}^{\text{bags(seg-back)}} \nonumber \\
& \cup \mathcal{P}_{\text{watch,read,lookup}}^{\text{bags(dict-back)}}.
\end{align}

\noindent To counterbalance the positives, we use $\mathcal{S}$ in combination with $\mathcal{M}$ and $\mathcal{D}$ to create four kinds of negative bags. Differently to positive sampling, negatives can be constructed across the full minibatch rather than solely from the current (subtitled video, dictionary) pairing. We first anchor negatives bags on foreground segments: 

\begin{align}
&  \mathcal{N}_{\text{watch,read,lookup}}^{\text{bags(seg-fore)}} = \{\{i\} \times B_i: (x_i^\mathcal{M}, v_i^\mathcal{M}) \in \mathcal{M}, \nonumber \\ 
& \quad \quad \quad \quad \quad \quad (x_j^\mathcal{D}, v_j^\mathcal{D}) \in \mathcal{D}, B_i = \{j : v_j^\mathcal{D} \neq v_i^\mathcal{M} \}\}
\end{align}

\noindent so that they contain pairs between a given foreground segment and all available dictionary videos whose label does not match the segment (visualised in Fig.~\ref{app:fig:sampling_mil2nce} (i), both rows). We next anchor on the foreground dictionary videos:

\begin{align}
    & \mathcal{N}_{\text{watch,read,lookup}}^{\text{bags(dict-fore)}} = \{A_i \times B_i:  (x_i^\mathcal{D}, v_i^\mathcal{D}) \in \mathcal{D}, \nonumber \\
    & \quad \quad \quad A_i = \{j : v_i^\mathcal{D} \in \text{Tokenize}(s_k), (x_k, s_k) \in \mathcal{S}, x_j \subseteq x_k, \nonumber \\
    & \quad \quad \quad (x_j, v_j) \notin \mathcal{M}\} \cup \{(x_m, v_m) \in \mathcal{M}, v_m \neq v_i \}, \nonumber \\
    & \quad \quad \quad B_i = \{l: v_l^\mathcal{D} = v_i^\mathcal{D}\}\}
\end{align}

\noindent comprising of pairings between the dictionary foreground set and segments within the minibatch that are either labelled with a different word, or can be excluded as a potential match through the subtitles (Fig.~\ref{app:fig:sampling_mil2nce} (ii), both rows).  Next, we anchor on the background continuous segments:
\nopagebreak
\begin{align}
& \mathcal{N}_{\text{watch,read,lookup}}^{\text{bags(seg-back)}} = \{\{i\} \times B_i: \exists (x_k, s_k) \in \mathcal{S}, x_i \subseteq x_k, \nonumber \\
& \quad \quad \quad (x_j^\mathcal{D}, v_j^\mathcal{D}) \in \mathcal{D}, 
 B_i = \{j : v_j^\mathcal{D} \notin \text{Tokenize}(s_k) \}\}
\end{align}

\noindent which amounts to the pairings between each background segment and the set of dictionary videos which do not correspond to any of the words in the background subtitles (Fig.~\ref{app:fig:sampling_mil2nce} (iii), both rows). The fourth negative bag set construction anchors on the background dictionaries:

\begin{align}
    \mathcal{N}_{\text{watch,read,lookup}}^{\text{bags(dict-back)}} = \{A_i \times B_i: (x_i^\mathcal{D}, v_i^\mathcal{D}) \in \mathcal{D}, \nonumber \\ A_i = \{j : v_i^\mathcal{D} \notin \text{Tokenize}(s_k), (x_k, s_k) \in \mathcal{S}, \nonumber \\ x_j \subseteq x_k, (x_j, v_j) \notin \mathcal{M}\} \cup \{(x_m, v_m) \in \mathcal{M}, v_m \neq v_i \}, \nonumber \\ B_i = \{l: v_l^\mathcal{D} = v_i^\mathcal{D}\}\}
\end{align}

\noindent and thus the pairings arise between dictionary examples for a background segment and its corresponding foreground segment, as well all segments from other batch elements (Fig.~\ref{app:fig:sampling_mil2nce} (iv), both rows).  These four sets of bags are combined to form the full negative bag set:

\begin{align}
&\mathcal{N}_{\text{watch,read,lookup}}^{\text{bags}} =
\mathcal{N}_{\text{watch,read,lookup}}^{\text{bags(seg-fore)}}
\cup \mathcal{N}_{\text{watch,read,lookup}}^{\text{bags(seg-dict)}} \nonumber \\ 
& \quad \quad \quad \quad \quad \cup \, \mathcal{N}_{\text{watch,read,lookup}}^{\text{bags(seg-back)}}
\cup \mathcal{N}_{\text{watch,read,lookup}}^{\text{bags(dict-back)}}.
\end{align}

\noindent In the main paper, these bag formulations are used through Eqn. (1) (the MIL-NCE loss function) to guide learning.  Concretely, the \textit{Watch-Lookup} framework defines positive and negative bags via $\mathcal{P}^{\text{bags}} =\mathcal{P}_{\text{watch,lookup}}^{\text{bags}}$, $\mathcal{N}^{\text{bags}} =\mathcal{N}_{\text{watch,lookup}}^{\text{bags}}$ and the \textit{Watch-Read-Lookup} formulation instead defines the positive and negative bags via  $\mathcal{P}^{\text{bags}} =\mathcal{P}_{\text{watch,read,lookup}}^{\text{bags}}$, $\mathcal{N}^{\text{bags}} =\mathcal{N}_{\text{watch,read,lookup}}^{\text{bags}}$.

\subsection{Infrastructure and runtime}\label{app:subsec:infra}

\noindent\textbf{Training.} The I3D trunk \bslonek{} pretraining experiments were performed with four Nvidia M40 graphics cards and took 2-3 days to complete. After freezing the I3D trunk, training the parameters of the MLP with the \textit{Watch-Read-Lookup} framework took approximately two hours on a single Nvidia M40 graphics card.

\noindent\textbf{Inference.} Our sign spotting demo available online (link at our project page) runs at real time in case of GPU availability. A single forward pass from the I3D and MLP layers takes 0.016 seconds to process 16 video frames on a single Nvidia M40 GPU, which is roughly 1000 frames per second (much more than the 25 fps real time capture speed). However, our current models (both for spotting and recognition) rely on the I3D model, which is a 3D convolutional neural network with about 15M parameters. Future work can focus on compressing these heavy models into more lightweight architectures for mobile applications.

\end{document}